\documentclass[lettersize,journal]{IEEEtran}
\usepackage{amsmath,amsfonts}
\usepackage{array}
\usepackage[caption=false,font=footnotesize]{subfig} 
\usepackage{textcomp}
\usepackage{stfloats}
\usepackage{url}
\usepackage{verbatim}
\usepackage{graphicx}
\usepackage{cite}
\usepackage{amsfonts} 
\usepackage[ruled]{algorithm2e}
\usepackage{booktabs}
\usepackage{hyperref}

\begin{document}

\title{HEADER: Hierarchical Robot Exploration via Attention-Based Deep Reinforcement Learning with Expert-Guided Reward}

\author{Yuhong Cao$^{1}$, Yizhuo Wang$^{1}$, Jingsong Liang$^{1}$, Shuhao Liao$^{1}$, Yifeng Zhang$^{1}$, Peizhuo Li$^{1}$, Guillaume Sartoretti$^{1}$
\thanks{$^{1}$ Authors are with the Department of Mechanical Engineering, College of Design and Engineering, National University of Singapore.}}

\maketitle

\begin{abstract}

This work pushes the boundaries of learning-based methods in autonomous robot exploration in terms of environmental scale and exploration efficiency.
We present HEADER, an attention-based reinforcement learning approach with hierarchical graphs for efficient exploration in large-scale environments.
HEADER follows existing conventional methods to construct hierarchical representations for the robot belief/map, but further designs a novel community-based algorithm to construct and update a global graph, which remains fully incremental, shape-adaptive, and operates with linear complexity.
Building upon attention-based networks, our planner finely reasons about the nearby belief within the local range while coarsely leveraging distant information at the global scale, enabling next-best-viewpoint decisions that consider multi-scale spatial dependencies.
Beyond novel map representation, we introduce a parameter-free privileged reward that significantly improves model performance and produces near-optimal exploration behaviors, by avoiding training objective bias caused by handcrafted reward shaping. 
In simulated challenging, large-scale exploration scenarios, HEADER demonstrates better scalability than most existing learning and non-learning methods, while achieving a significant improvement in exploration efficiency (up to 20\%) over state-of-the-art baselines.
We also deploy HEADER on hardware and validate it in complex, large-scale real-life scenarios, including a $300m\times230m$ campus environment.
\end{abstract}


\section{Introduction}
\IEEEPARstart{I}{n} autonomous exploration, a mobile robot is tasked with exploring and mapping an unknown environment as fast as possible. By planning and executing its exploration path, the robot classifies \textit{unknown} areas into \textit{free} or \textit{obstacle} areas based on its accumulated sensor measurements. In this work, we focus on tasks where a ground robot is equipped with an omnidirectional 3D LiDAR to obtain long-range, low-noise, and dense point cloud measurements. Recent advancements in LiDAR odometry have enabled accurate and robust localization and mapping in large-scale environments~\cite{zhang2014loam,shan2020lio,xu2022fast}, allowing recent planners to focus on exploring the environment without concerns about mapping/localization accuracy~\cite{selin2019efficient,cao2021tare,dang2020graph,huang2023fael,peltzer2022fig,long2024hphs}. Despite this, few planners support exploration at large scale in real-world environments~\cite{cao2021tare,cao2023representation}, mainly due to the complexity that comes with long-term, real-time path planning requirements. That is, to achieve efficient exploration, the planner must actively react to belief and map updates at a high frequency by (re-)reasoning about the full partial belief, to replan a long-term, non-myopic exploration path. For example, in one of our simulation environments, our robot needs to explore a $330m\times250m$ environment, where the robot needs to reason about up to $10,000 m^2$ areas at a resolution of $1.2m$ and a frequency of $2Hz$.

\begin{figure}[t]
    \centering
    \includegraphics[width=0.45\textwidth]{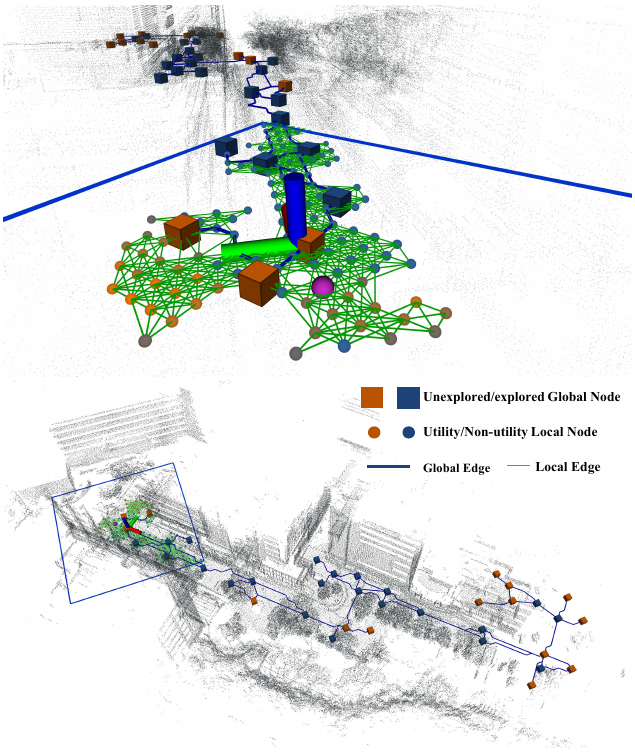}
    \caption{Example hierarchical graph constructed by HEADER during its autonomous exploration of our campus. While dense local nodes keep expanding to cover all traversable areas, our planner adaptively and incrementally partitions these local nodes (bounded by the blue box as the local range) into global nodes to generate the global reference path. Based on the local graphs and the global path, our trained model selects a neighboring node as the next waypoint (denoted as the purple ball) to visit.}
    \label{fig:intro}
    
\end{figure}

Reasoning about such a large-scale map at full resolution is non-trivial when a high replanning frequency is desired.
Therefore, most recent advanced large-scale exploration planners rely on hierarchical planning~\cite{selin2019efficient,cao2021tare,peltzer2022fig,zhu2021dsvp,long2024hphs}, which decouples the exploration path into a low-resolution global path with coarse inference on the full belief and a high-resolution local path with fine reasoning on the nearby local belief.
The insight of hierarchical planning is that fine reasoning is most effective in nearby areas, while less detailed inference is sufficient for distant utility.
We note that a sparser global representation is the key to better scalability, since it leads to lower complexity for global planning.
Existing methods~\cite{cao2021tare, cao2023representation, long2024hphs} that rely on predefining fixed global resolutions currently demonstrate the best scalability. However, the choice of global resolutions is crucial, and may require further tuning for different environments: high resolution will reduce planning efficiency, while low resolution risks losing paths to specific areas and thus failing to complete the task. Intuitively, a shape-adaptive but still fully incremental method should work best to represent the global map.

Learning-based methods are intuitively promising to tackle autonomous exploration since they take advantage of predicting future states (i.e., unknown areas) based on previous experience, instead of only relying on current states (i.e., the current free areas) like conventional planners.
However, existing learning-based works have yet to explore how to integrate with hierarchical planning, since most existing learning-based planners still struggle to achieve good performance (i.e., exploration rate, makespan, and sim-to-real robustness) in small/local-scale exploration~\cite{chen2019self,chen2020autonomous,xu2022explore,zhu2024maexp,niroui2019deep}.
Notably, Cao et al. recently introduced ARiADNE~\cite{cao2023ariadne}, which proposed attention decision networks over graph-based representations to better capture both short- and long-range dependencies in autonomous exploration.
Compared to previous convolutional neural network-based planners~\cite{chen2019self,chen2020autonomous,xu2022explore,zhu2024maexp}, ARiADNE achieves efficient exploration in small-scale exploration tasks and provides robust sim-to-real ability.
However, in more complex and larger-scale scenarios, ARiADNE still suffers from the sparse reward problem.  
Sparse reward is a general but critical challenge for deep reinforcement learning (DRL)~\cite{devidze2022exploration,trott2019keeping,memarian2021self}, which will cause vanishing gradients and sample inefficiency.
More specifically, in autonomous exploration, the only relevant metric is the distance/time to complete the task, but rewarding the robot only at the end of an episode/task may make the reward signal too sparse to learn.
Existing learning-based exploration methods often tackled this issue through handcrafted reward shaping~\cite{chen2019self,chen2020autonomous,xu2022explore,zhu2024maexp}, where the robot receives a dense reward (e.g., proportional to the observed information) at each time step.
However, this local incentive may compromise the global objective of minimizing the total exploration path.
Moreover, such reward shaping still fails in environments that require long-term backtracking (e.g., long corridors), where no information is observable for prolonged periods, even as the robot acts optimally.
As a result, the scalability and exploration efficiency of existing learning-based planners are still far from matching that of conventional state-of-the-art methods.

In this work, we push the boundary of learning-based robot exploration by proposing efficient hierarchical graphs and tackling the reward bias issue, making our planner the first learning-based approach that achieves better scalability and exploration performance than state-of-the-art conventional planners.
We introduce HEADER, which substantially extends ARiADNE by: (1) enabling powerful scalability through a novel shape-adaptive and fully incremental global graph based on community detection~\cite{traag2019louvain}; (2) enhancing the model's long-term reasoning ability by designing novel guidepost features; (3) significantly improving its exploration performance through a privileged expert reward training.
We first propose a global graph construction method to enable efficient global planning, which can incrementally generate shape-adaptive subregions for arbitrary structures without parameter tuning.
Our method leverages community detection~\cite{traag2019louvain} to adaptively partition the dense graph into global subregions by identifying clusters of nodes with high intra-community connectivity and structural cohesiveness~\cite{fortunato2010community}.
Then, we inject global information into local observations by designing guidepost features as our network inputs, so our planner can make decisions on local movements (waypoint a few meters away) while optimizing the performance of whole-environment exploration.
Our guidepost features are binary signals that indicate whether local nodes are on local or global reference paths leading to areas yet to be explored.
The intuition is that for distant pockets of information, what truly matters is merely being aware of their existence and knowing how to reach them, while details like their distribution/shape might be omitted.
By doing so, we avoid the burden of training the model for long-term pathfinding and allow the model to focus on discovering dependencies between areas to be explored.  
Last but not least, to overcome reward shaping biases, we introduce a parameter-free privileged expert reward that allows the robot to optimize the true task objective.
At each training step, by allowing an expert planner to access the full environment, we compute an optimal coverage path from the current position.
We then calculate a privileged reward based on the Euclidean distance between the privileged expert action and the action our planner outputs, which constantly provides dense and unbiased feedback for actions in any kind of environment.
This design provides a precise evaluation of each action's impact on the full exploration process, thereby allowing HEADER to output decisions that are more likely to benefit the exploration efficiency.

Through these novel contributions, HEADER demonstrates state-of-the-art exploration efficiency (on average $20\%$ higher than the second-best planner~\cite{cao2021tare} in the benchmark tests) and scalability (completing exploration in large-scale environments where most recent open-sourced planners~\cite{dang2020graph,huang2023fael,long2024hphs} fail). We also developed a terrain segmentation tool based on Octomap~\cite{hornung13auro} to enable the deployment of HEADER in 3D environments. We validated HEADER in multiple indoor and outdoor simulation benchmarks, as well as on hardware in real-life scenarios, including a $300m\times230m$ outdoor environment, where HEADER guided a wheeled robot over $1.5$ kilometers to explore over $60,000m^3$ of our campus.
To the best of our knowledge, this is the first successful demonstration of applying a learning-based exploration planner in real-world tests at such a scale.
Finally, our ablation tests quantify that HEADER achieves up to $50\%$ accuracy in implicitly predicting unknown areas that are key to exploration decisions, which we believe allows us to start probing into the underlying mechanisms explaining the advantage of HEADER over conventional planners.

\section{Related Works}

\subsection{Conventional Exploration}
While a group of works formulate the exploration problem based on information theory~\cite{zhang2020fsmi,asgharivaskasi2022active}, most conventional exploration planners can be categorized as frontier-driven methods, which are designed to identify and navigate to \textit{frontiers} (i.e., the boundaries between known free and unknown areas) to expand the robot’s belief of the environment. Early exploration methods directly select a frontier point as the navigation goal~\cite{yamauchi1997frontier,gonzalez2002navigation}, which leads to greedy and myopic performance. Subsequent works reformulate exploration as a viewpoint selection problem~\cite{bircher2016receding}, where the exploration path is considered as a sequence of viewpoints that maximize the \textit{utility} (i.e., expected new explored areas). While such formulation allows non-myopic long-term planning and careful consideration of occlusion, it inevitably incurs high computational complexity in this real-time task. As a result, sampling-based approaches with receding horizon have emerged as the mainstream strategy for autonomous robot exploration. Bircher et al.~\cite{bircher2016receding} first proposed to leverage a rapidly-exploring random tree (RRT) to sample exploration paths. Selin et al.~\cite{selin2019efficient} further improved~\cite {bircher2016receding} by optimizing the sampling strategy through cached points and Gaussian Process. To ensure complete coverage of traversable areas, more recent studies have shifted towards graph-based sampling~\cite{dang2020graph,cao2021tare,xu2021autonomous,huang2023fael}. Dang et al.~\cite{dang2020graph} proposed to construct a rapidly-exploring random graph (RRG) and iteratively sample exploration paths. While~\cite{dang2020graph,xu2021autonomous} adopt strategies from~\cite{bircher2016receding} to sample feasible paths, Cao et al.~\cite{cao2021tare} introduced a combination of constrained sampling and the Traveling Salesman Problem (TSP) to efficiently improve frontier coverage. Inspired by TSP solvers, Huang et al.~\cite{huang2023fael} proposed a viewpoint sequencing method based on the 2-opt local search heuristic.

Among them, many recent works~\cite{selin2019efficient,dang2020graph,zhu2021dsvp,cao2021tare,huang2023fael,long2024hphs} recognize the importance of hierarchical planning in enhancing the scalability of autonomous robot exploration. A common strategy is to employ a high-resolution, sampling-based approach for fine-grained viewpoint selection at the local scale, combined with a low-resolution, frontier-based method for generating coarse yet long-range paths at the global scale. However, designing a general global representation of the map that maintains effectiveness in different environments is non-trivial. The major challenge is the trade-off between representation sparsity and fine coverage of the map. A group of methods extracts global representations from local representations. Selin et al.~\cite{selin2019efficient} directly used the frontiers as the global representation and relied on naive frontier-based planning. Dang et al.~\cite{dang2020graph} constructed a sparse global graph by adding high-gain paths from the local graph. Zhu et al.~\cite{zhu2021dsvp} improves the coverage of the global graph to deal with environments with more complex structures. These methods usually struggle to maintain sparsity in larger-scale autonomous exploration. Another group of methods predefines a global resolution to uniformly decompose the environment into subregions. Cao et al.~\cite{cao2021tare} decomposed the environment into subregions and planned the global path by finding the TSP solution over all unexplored subregions. Long et al.~\cite{long2024hphs} generated subregions in the same way, but planned global paths through maximizing a gain function. Although predefined resolutions improve scalability in large-scale environments by enabling sparser global representations, they require environment-specific tuning (e.g., whether there are many thin and long corridors) to balance coverage and sparsity\cite{cao2021tare,cao2023representation}, which is impractical given that the environment structure is unknown. 

\vspace{-0.2cm}
\subsection{Community Detection}

Similar to clustering (e.g., k-means~\cite{hartigan1979algorithm}), community detection~\cite{fortunato2010community} is used to partition a group of nodes, but based on the graph structure: it aims to identify groups of nodes within a graph that are more densely connected internally than sparsely connected externally. While clustering methods usually assume roughly balanced cluster sizes, community detection naturally identifies groups with significant size differences, making it more suitable for exploration tasks. Besides, since general clustering methods do not consider the connectivity of nodes, they can not recognize some spatial structures (e.g., two adjacent rooms separated by a wall) as effectively as community detection. Moreover, community detection can also automatically determine the number of partitions. 
Newman~\cite{newman2006modularity} pioneered modularity-based community detection and developed a greedy algorithm to optimize the modularity of the network partitions. Blondel et al.~\cite{blondel2008fast} proposed the Louvain algorithm to improve computational efficiency in large-scale graphs. Traag et al.~\cite{traag2019louvain} further improve the modularity optimization process of~\cite{blondel2008fast}, offering faster and more accurate community detection. We note that community detection is a desired tool to extract global representation in autonomous exploration. It can sparsely partition environments with arbitrary shapes with low complexity, which is non-trivial for partition methods applied in existing works. For example, the Voronoi-based method~\cite{hu2020voronoi} can only output convex subregions, thus resulting in dense partitions in challenging environments; the skeleton-based~\cite{noel2023skeleton} method needs per-pixel operation, leading to a high computing burden. Besides, those methods are not naturally incremental.

\subsection{Learning-based Exploration}

There is a group of works that focus on explicit map prediction through supervised learning~\cite{luperto2021exploration,georgakis2022uncertainty,ho2024mapex}, which predict the layout of structured indoor environments and integrate with conventional frontier-based planners. Those methods likely will suffer from poor out-of-distribution generalizability. More importantly, the overall exploration performance is still constrained by conventional exploration planners. More works have tried to improve the exploration performance through training learning-based planners to make decisions~\cite{niroui2019deep,chen2019self,chen2020autonomous,xu2022explore,cao2023ariadne,zhu2024maexp,liang2024hdplanner}. Those works leverage deep reinforcement learning to train neural networks to estimate the long-term return and the optimal exploration policies. Many of them are based on convolutional neural networks (CNNs)~\cite{niroui2019deep,chen2019self,xu2022explore,zhu2024maexp}, while choosing different action space:~\cite{niroui2019deep,xu2022explore} select frontiers to visit;~\cite{zhu2024maexp} directly outputs acceleration
and steering angle;~\cite{chen2019self} chooses nearby candidate viewpoints. A major limitation of CNN-based planners is that the input dimension of the network must remain fixed, which restricts their flexibility in handling environments of varying sizes/resolutions. Besides, in real-world applications, the constructed maps often contain noise, which degenerates the generalizability of CNN-based planners. Cao et al.~\cite{cao2023ariadne,cao2024deep} demonstrate that graph learning-based methods are promising in addressing these challenges. By formulating exploration as a next viewpoint selection problem on a graph and leveraging attention-based neural networks for graph processing, ~\cite{cao2023ariadne} can naturally handle varying environment sizes. The graph representation also helps to filter out mapping noise and recognize underlying exploration essentials, enabling robust sim-to-real transfer as shown in~\cite{cao2024deep,tan2024ir}.
Advanced hierarchical planning is rarely studied in learning-based methods. Chen et al.~\cite{chen2019self} proposed to use the naive frontier-based method to reposition the robot when there is no frontier in the range of their CNN networks. Liang et al.~\cite{liang2024hdplanner} designed a hierarchical strategy to select frontiers and viewpoints sequentially to make joint decisions. Although these works incorporate hierarchical designs, they do not leverage advanced global representations to enhance scalability, as achieved by state-of-the-art conventional planners.

\subsection{Privileged Learning and Imitation Learning}

Inspired by human learning, where teachers provide additional insights beyond the standard examples, privileged learning provides additional/privileged information to the model during training but not during testing~\cite{vapnik2009new}. It aims to improve generalization and performance by leveraging the extra information. Although the concept of privileged learning was initially proposed for conventional machine learning~\cite{vapnik2009new,vapnik2015learning}, the idea to leverage privileged information has been widely adopted in robot learning~\cite{loquercio2021learning,lee2020learning,monaci2022dipcan,bajcsy2024learning}. Loquercio et al.~\cite{loquercio2021learning} leveraged privileged learning for outdoor UAV navigation, where they generated privileged expert paths and trained the policy network through imitation learning. Lee et al.~\cite{lee2020learning} introduced privileged learning for quadrupedal locomotion, where they train a teacher network with privileged observation and let a student policy imitate the teacher policy. Similarly, Monac et al.~\cite{monaci2022dipcan} utilized privileged learning and imitation learning for crowd-aware navigation, where they adapt non-privileged information to replace the privileged input. 

It could be seen that privileged learning and imitation learning are tightly coupled. The above methods were integrated with behavior cloning, where the training objective is to minimize the difference between the policy output and the expert action. Behavior cloning is a naive imitation learning method with known drawbacks, particularly its lack of exploration in the policy space, which often results in poor generalization to out-of-distribution scenarios compared to reinforcement learning. This challenge is more severe in exploration tasks compared to locomotion and navigation. A more advanced imitation learning method is inverse imitation learning~\cite{wulfmeier2015maximum}, which aims to learn the reward function for DRL from expert demonstrations to enable exploration in the expert policy space. Another notable effort to address this challenge is generative adversarial imitation learning~\cite{ho2016generative}, which trains a discriminator to distinguish between expert trajectories and those generated by the network. The adversarial training strategy can also be utilized to learn the reward function~\cite{fu2017learning}. However, these methods rely on complex networks or training processes, making them hard to apply in many tasks. A simpler yet efficient approach is imitation reward~\cite{peng2018deepmimic}, which demonstrates strong performance in locomotion learning by defining the reward as the difference between the learned joint velocities/positions and the expert references.

\section{Problem Formulation}

\begin{figure*}[t]
    \centering
    \includegraphics[width=0.95\linewidth]{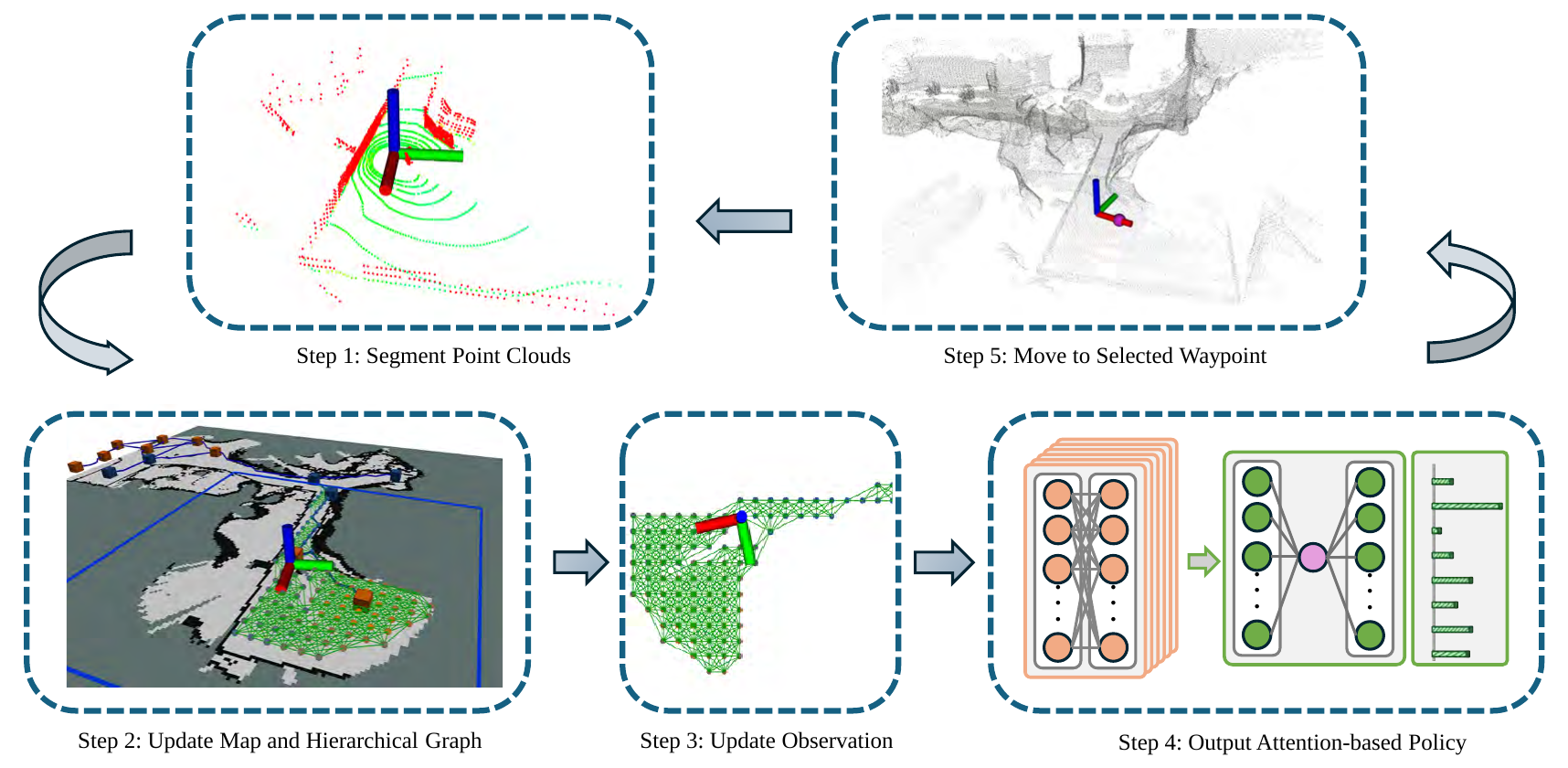}
    \caption{Overall working flow of HEADER during deployment. We first classify the sensor scan into free and obstacle point clouds, which are used to build the current map of the environment. Then, we construct the community-based hierarchical graph to generate the graph observation as the input of the trained neural network. After that, our attention-based decision model outputs the next waypoint for the robot to follow and get new measurements.}
    \label{fig:main}
\end{figure*}

Let $\mathcal{M} \subset \mathbb{R}^3$ be the current map/belief of the environment to be explored. Define $\mathcal{M}_{free} \subseteq \mathcal{M}$ as the known free/traversable space, $\mathcal{M}_{obs} \subset \mathcal{M}$ as the known obstacle/occupied space, and $\mathcal{M}_{unk} \subset \mathcal{M}$ as the unknown space. We then define $\mathcal{M}_{fron} \subset \mathcal{M}$ as the frontier, the boundary between free space and unknown space. At each decision step $t$, the ground mobile robot plans and visits the next waypoint $w_t \in \mathcal{M}_{free}$ in the traversable space to observe the environment. The newly perceived unknown space by an omnidirectional Lidar with a sensor range $d_{sensor}$, will be classified as free space or occupied space based on the traversability. We denote the executed trajectory as $\psi=[p_0, p_1, ..., p_t], \forall p_i \in \mathcal{M}_{free}$, where $p_i$ denotes the position visited by the robot. We consider the exploration tasks as completed when no frontiers are remaining, such that $\mathcal{M}_{fron}=\emptyset$. The objective is to minimize the cost of the executed path to complete the exploration:
\begin{equation}
\label{eq: objective}
    \psi^*= \mathop{\rm {argmin}} \limits_{\psi \in \Psi}{\rm C}(\psi),\ {\rm {s.t.}}\ \mathcal{M}_{fron}=\emptyset,
\end{equation}
where $\psi^*$ is the optimal exploration path, $\rm C: \psi \xrightarrow{} \mathbb{R}^+$ maps a trajectory to its length, and $\Psi$ is the space of all feasible exploration paths.

\section{Methodology}

\begin{figure*}
\centering

\subfloat[Global Graph\label{fig:global_graph}]{
    \includegraphics[width=0.24\textwidth]{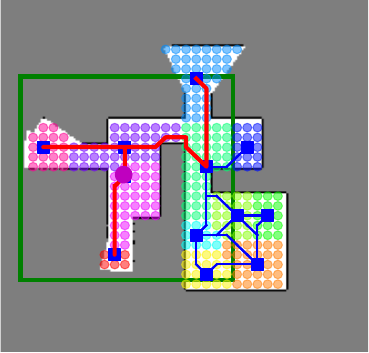}
}
\hfill
\subfloat[Node Utility\label{fig:node_utility}]{
    \includegraphics[width=0.23\textwidth]{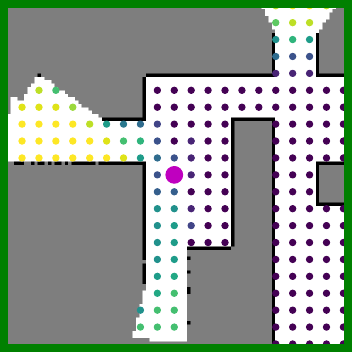}
}
\hfill
\subfloat[Local Guidepost\label{fig:local_guidepost}]{
    \includegraphics[width=0.23\textwidth]{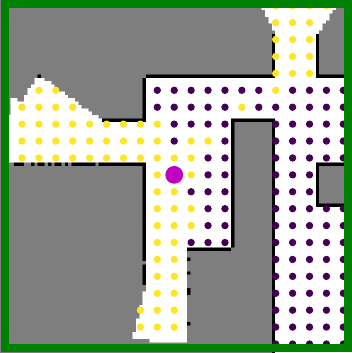}
}
\hfill
\subfloat[Global Guidepost\label{fig:global_guidepost}]{
    \includegraphics[width=0.23\textwidth]{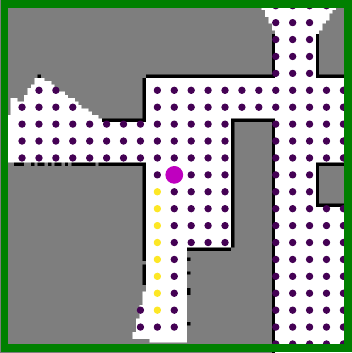}
}

\caption{Example community-based hierarchical graph constructed by HEADER. The purple dot indicates the robot's current position, and the green square the local graph region. Small dots denote local nodes, and blue squares global nodes. The blue paths denote the global edges, and the red path the planned global path. In Fig.~\ref{fig:global_graph}, the colors of local nodes represent their community memberships, while in Fig.~\ref{fig:node_utility}, they reflect node utilities. Fig.~\ref{fig:local_guidepost} and~\ref{fig:global_guidepost} highlight the selected local and global guideposts, respectively. Local guideposts denote paths to all utility nodes, and global guideposts paths to the next global node.}

\label{fig:graph}
\end{figure*}

\begin{figure}[b]
    \centering
    \includegraphics[width=\columnwidth]{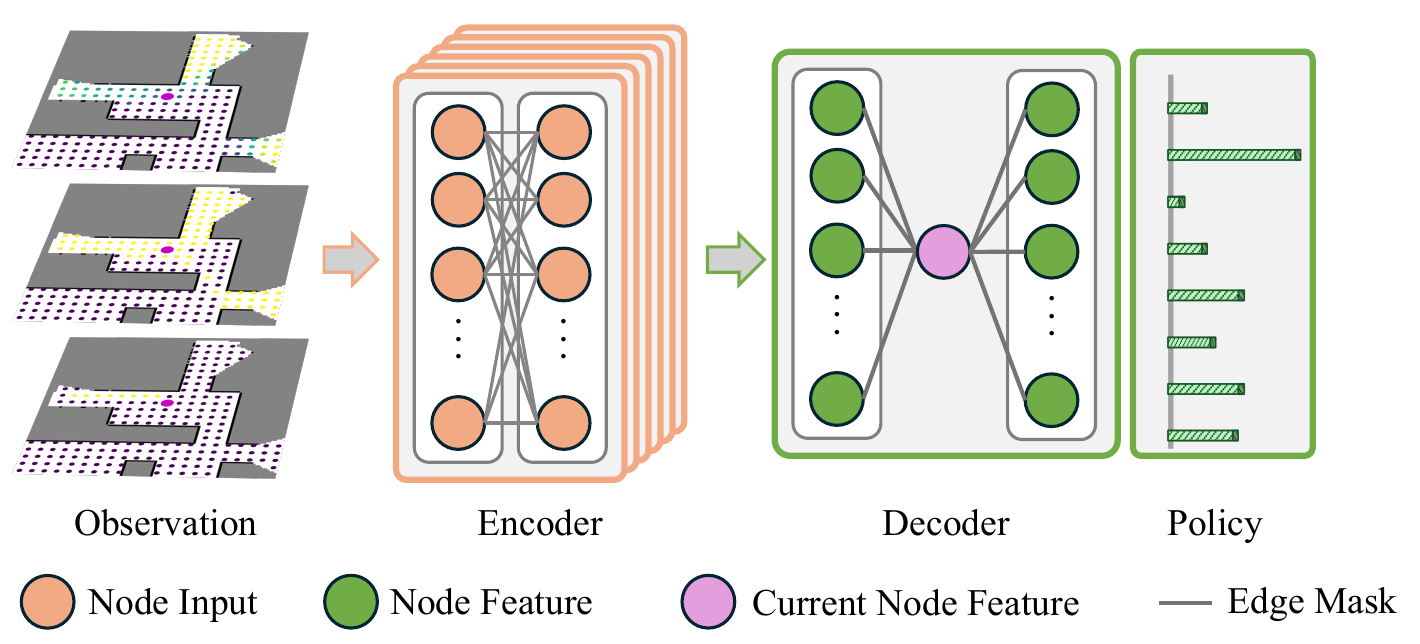}
    \caption{Attention-based decision network in HEADER. It takes the local graph, utility, and guideposts as input and outputs a probability distribution over neighboring nodes of the current robot position as its policy.}
    \label{fig:network}
\end{figure}

In this section, we detail the working flow of HEADER (see Fig.~\ref{fig:main}) and our training paradigm. We first uniformly distribute candidate viewpoints in the free space and connect each viewpoint with its collision-free neighbors to construct a local graph that densely covers the current traversable space. We then run our community detection algorithm to incrementally and adaptively partition the local graph to construct the global graph (see Fig.~\ref{fig:global_graph}). Based on the global graph, we generate a global reference path by solving a TSP problem to visit all unexplored global nodes. After that, we formulate an informative graph that enhances the nearby local graph with frontier information (see Fig.~\ref{fig:node_utility}) and reference paths (see Fig.~\ref{fig:global_guidepost} and~\ref{fig:local_guidepost}). Finally, we input this informative graph into our attention-based network (see Fig.~\ref{fig:network}), which selects one of the neighboring viewpoints as the next waypoint to visit.

\subsection{Community-based Hierarchical Graph}

\subsubsection{Local Graph}
We first construct a \textit{full dense graph} $G_d=(V_d,E_d)$ by incrementally and uniformly adding candidate viewpoints/waypoints at node resolution $\Delta_{node}$ in the free space. The node set is defined as $V_d=\{v_1,v_2,..,v_n\}, \forall v_i=(x_i,y_i) \in \mathcal{M}_{free}$. For each viewpoint $v_i$, we identify its neighboring viewpoints through line-of-sight checking (i.e., the distance between two viewpoints is within a threshold $d_n$ and the straight line between them lies entirely in the free space). The neighboring connections formulate a set of collision-free edges $E_d=\{(v_1,v_2),...,(v_{n-1},v_n)\}$. At each decision step $t$, we define the \textit{robot node} $v_{cur}$ as the nearest node to the robot's current position $p_t$. Then we define a $d_{local}\times d_{local}$ sliding window $\mathcal{W}$ centered at $v_{cur}$ and construct the \textit{local graph} $G_l=(V_l, E_l)$, where the local node set $V_l \subseteq V_d$ contains all candidate nodes within $\mathcal{H}$, and the local edge set $E_l\subseteq E_d$ includes all edges between nodes in $V_l$. 

\subsubsection{Global Graph}
Leveraging the Leiden algorithm~\cite{traag2019louvain}, we adaptively partition the local graph $G_l$ to incrementally construct a sparse \textit{global graph} $G_g=(V_g,E_g)$, while ensuring full coverage of the robot belief (i.e., all local nodes remain reachable through the global graph). The Leiden algorithm aims to maximize the modularity of the partition, which is defined as:
\begin{equation}
    Q=\frac{1}{2m}\sum_{i,j}[A_{ij}-\beta\frac{k_ik_j}{2m}]\delta(c_i, c_j),
\end{equation}
where $A_{ij}$ is the adjacent matrix of the local graph, $k_i=\sum_jA_{ij}$ is the degree of node $i$, $m=\frac{1}{2}\sum_{i,j}A_{ij}$ is the total number of edges in the graph, $\beta$ is a linear resolution parameter, $c_i$ denotes the community that node $i$ belongs to, $\delta(c_i, c_j)$ is an indicator function, which equals $1$ if nodes $i$ and $j$ belong to the same community and $0$ otherwise. Modularity evaluates whether the partition result is more structured than the random partition. High modularity indicates a community structure with dense intra-community and sparse inter-community connections.

Starting with an initial partition $\{C_1, C_2, ...,C_k\}$, for each node $i$, we move it from the current community to its neighboring community (i.e., the community that its neighboring node belong to and not identical to the current community), if modularity gain $\Delta Q>0$. This node movement process continues until the modularity stops increasing. We then refine the community partition to ensure that nodes in each community are internally connected. We perform the connectivity check based on $E_l$ to split communities that are not fully internally connected. Subsequently, we merge the subcommunities with their neighbors if the modularity can be improved by doing so. To ensure consistency of the partition results, we fix the community assignments of existing nodes and only perform community assignments for newly added nodes at the current decision step. We also set a threshold based on $\mathcal{W}$ and $\Delta_{node}$ to limit the maximum number of nodes allowed in each community. To formulate $G_g$, for each community, we generate a global node $g_i\in V_g$ at the nearest local node to the center of the community and find its path and cost to neighboring global nodes by running A*~\cite{hart1968formal} on the local graph. 
We term the global node corresponding to the robot node's community as the \textit{current global node} $g_{cur}$ and relocate its position to the same position as the robot node $v_{cur}$. 

\subsubsection{Incorporating Frontiers} 
Now we incorporate frontier information into the hierarchical graph. For each local node $v_i$, we compute its \textit{utility} $u_i$, which represents the number of observable frontiers from the position $v_i$ located within a utility range $d_{utility}$ (smaller than the sensor range). We classify the local nodes into \textit{utility nodes} and \textit{non-utility nodes}, depending on whether their utility value satisfies $u_i>0$. A global node is classified as an \textit{unexplored global node} or an \textit{explored global node}, depending on whether it contains at least one utility node.

\subsubsection{Reference Paths}
We compute two types of reference paths: a \textit{global reference path} $P_g$ and \textit{local reference paths} $P_l$ for the following decision-making. The global reference path is the shortest path to visit all unexplored global nodes starting from $g_{cur}$. We obtain this path by solving a TSP problem using OR-Tools~\cite{ortools}. Each local reference path corresponds to the shortest path from $v_{cur}$ to a utility node. We obtain these paths using Dijkstra's algorithm~\cite{dijkstra1959note}. In all cases, the initial portion of the current optimal exploration path lies within the set of these reference paths.

\subsection{Attention-based Decision}

\subsubsection{Observation}

We combine the local graph, utility, and reference paths to formulate the observation $o_t=(G'_l, p_t)$, which serves as the network inputs. We term the combined results as the \textit{informative graph} $G_l'=(V_l', E_l)$, which shares the same edge set with the local graph $G_l'$ and the nodes' positions are identical to the local node set $V_l$. In addition to the position $(x_i, y_i)$, each node $v'_i$ in the informative graph further contains the utility $u_i$, local guidepost $e_i$, and global guidepost $b_i$, such that $v'_i=(x_i,y_i,u_i,e_i,b_i),\forall v'_i\in V'_l$. A global/local guidepost is a binary signal that denotes whether the location of the node is in the set of global/local reference paths. Note that the global graph is not included in the observation, as its information has already been exploited through the global reference path.

\subsubsection{Action} 
At each decision step $t$, our planner selects a neighboring node $v_i$ of $v_{cur}$ as the next waypoint $w_t$. The robot then executes action $a_t$, which navigates it to $w_t$. The policy is denoted as $\pi_\theta(a_t|o_t)=p_\theta(v_i,(v_i,v_{cur})\in E_l|o_t)$, where $\theta$ represents the set of weights of the policy network, $p$ is a discrete probability function over neighboring nodes of the current robot node $v_{cur}$.

\subsubsection{Policy Network}
Our policy network consists of an \textit{encoder} and a \textit{decoder}, both of which are built upon attention layers~\cite{vaswani2017attention} as their fundamental components, allowing them to naturally handle the graph input. The attention layer reads:
\begin{equation}
\begin{aligned}
& \textbf{q}_i=W^q\textbf{h}^{q}_{i}, \ \textbf{k}_i=W^k\textbf{h}^{k,v}_{i}, \ \textbf{v}_i=W^v\textbf{h}^{k,v}_i, \ \textbf{u}_{ij}=\frac{\textbf{q}_{i}^T\cdot \textbf{k}_{j}}{\sqrt{d_{f}}}, \\ & \textbf{w}_{ij}=\left\{\begin{array}{cc}
    \frac{e^{\textbf{u}_{ij}}}{\sum_{j=1}^{n}e^{\textbf{u}_{ij}}}, & M_{ij}=0 \\
    0, & M_{ij}=1
\end{array}\right., \ \textbf{h}'_{i}=\sum_{j=1}^{n}\textbf{w}_{ij}\textbf{v}_{j},
\end{aligned}
\end{equation}
where each $\textbf{h}_i\in\mathbb{R}^{d_f\times 1}$ is a $d_f$-dimension feature vector projected from node $v'_i$, superscripts $\textbf{q}$ and $\textbf{k},\textbf{v}$ denote the source of the \textit{query}, \textit{key}, and \textit{value} respectively, $W^q, W^k, W^v \in \mathbb{R}^{d_f\times d_f}$ are learnable matrices, and $M$, which serves as an edge \textit{mask}, is an adjacency matrix built from $E_l$ ($M_{ij}=0$, if $(v_i, v_j) \in E_l$, else $M_{ij}=1$).

We use the encoder to model both short and long-range dependencies among $G'_l$, by allowing each node to selectively merge features from its neighbors. In the encoder, we stack six attention layers, each taking the output of its previous attention layer as input, where the query, key, and value source are the same: features of all nodes.
By constraining attention to direct neighboring nodes, we enable the model to precisely capture the connectivity of the graph. At the same time, the multi-layer architecture enables each node to aggregate information from distant nodes through multi-hop attention, thereby enhancing the model’s awareness of long-range dependencies. We term the output of the encoder as \textit{node features}, the feature corresponding to the robot node as \textit{current node feature}, and the feature corresponding to neighboring nodes as \textit{neighboring node features}.

We then use the decoder, which contains a cross-attention layer and a pointer layer, to output the final policy $\pi$.
The cross-attention layer selectively integrates relevant node features for the current decision by using the current node feature as the query and all node features as keys and values. We concatenate its output with the current node feature and project it back to a $d_f$-dimension feature, which serves as the final representation of the current state. The pointer layer then conducts attention weights calculation with this final feature as the query source and the neighboring node features as the key source. Specifically, the pointer layer directly outputs the attention weights, which can be considered as a probabilistic distribution $p$, as the policy $\pi$. Note that the dimension of the policy is determined by the number of neighboring nodes, allowing our policy network to handle arbitrary graph structures.

\subsection{Training with Privileged Expert Reward}

\begin{figure}[t]
    \centering
    \includegraphics[width=0.8\linewidth]{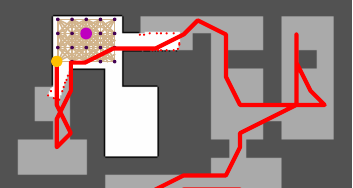}
    \caption{Illustration of the privileged expert path. The white area denotes the explored free space, while the light gray region indicates unexplored free space. The red trajectory represents the privileged expert path planned using ground-truth information. The yellow dot marks the expert path's first point as the expert waypoint. The light purple dot indicates the robot's current position, and the dark purple dots represent the candidate waypoints. Our privileged reward is based on the distance between the selected waypoint and the expert waypoint.}
    \label{fig:expert_reward}
\end{figure}

\subsubsection{Privileged Expert-guided Reward}

We first introduce our privileged expert planning algorithm, which is probabilistically complete to produce (near-)optimal coverage paths as demonstrations for training the policy network. We grant the planner access to the ground-truth environment, making it a privileged expert. By doing so, the partially observable exploration problem is transformed into a fully observable sensor coverage problem, which is easier to tackle as it eliminates the need to handle the uncertainty that lies in the unknown environment. We modify the frontiers coverage algorithm proposed in~\cite{cao2021tare} to plan a path that guarantees to cover all free space. Let $\mathcal{M}^*_{free}$ be the \textit{ground truth free space}, $\mathcal{M}^*_{obs}$ the \textit{ground truth obstacle space}. We denote the boundary between ground truth free space $\mathcal{M}^*_{free}$ and the unexplored obstacle space $\mathcal{M}^*_{obs}\setminus\mathcal{M}_{obs}$ as the \textit{privileged frontiers} $\mathcal{M}^*_{fron}$. We note that finding the shortest path to cover the entire ground truth free space is equivalent to covering all privileged frontiers. To this end, we first sample a set of viewpoints that collectively observe all frontiers, and then solve a TSP to compute the shortest path that visits each viewpoint. We iterate this process multiple times and select the path with the minimum travel distance as the privileged expert path. To maintain the consistency between the expert action space and the learned policy space, similar to the dense graph $G_d$, we construct a \textit{privileged graph} $G^*=(V^*, E^*)$, where $V^*\supseteq V_d, E^*\supseteq E_d$ as the planning space. We present the details in Algorithm~\ref{alg: ground truth planner}. Note that there is no hierarchical representation since we aim to ensure the optimality of the planned path for training.

\begin{algorithm}[t]
\caption{Privileged Expert Planner}
\label{alg: ground truth planner}
\KwIn {Privileged graph $G^*(V^*, E^*)$, current robot position $p_t$, privileged frontiers  $\mathcal{M}^*_{fron}$}
\KwOut {Privileged expert path $\psi^*_t$}
Get observable frontiers ${M}^*_{i,fron}$ for each node $v^*_i$ \\
Initialize priority queue $Q$ \\
$\forall v^*_i\in V^*$, push $v^*_i$ into $Q$ with priority as $u^*_i=|{M}^*_{i,fron}|$ \\
$C_{best} \gets +\infty$, $\psi^*_t \gets \emptyset$\\
\For{$1 \ \mathrm{to} \ k$}{
    $Q' \gets Q$, $P \gets \{p_t\}$ \\ 
    \While {$Q'\neq \emptyset \ \mathrm{and}\ \exists u^*_i>0$}{
        Probabilistically sample $ v^*_j \sim Q'$\\
        $Q' \gets Q'\setminus v^*_j$, $P \gets P\cup v^*_j$, \\
        ${M}^*_{fron}\gets \mathcal{M}^*_{fron}\setminus{M}^*_{i,fron}$ \\
        Update ${M}^*_{i,fron}$ and $u^*_i$ for all nodes in $Q'$ \\
        }
    Compute the TSP path $\psi$ over $P$ starting from the nearset node to $p_t$ \\
    \If {$C(\psi)<C_{best}$}{
        $C_{best} \gets C(\psi)$, $\psi^*_t \gets \psi$
    }
}
\end{algorithm}

At each decision step $t$, we compute the privileged expert path $\psi^*_t$ and take the next waypoint in $\psi^*_t$ as the expert waypoint $w^*_t$. We formulate the privileged expert reward $r_t$ based on the normalized Euclidean distance $d_t$ between the expert waypoint $w^*_t$ and the waypoint $w_t$ selected by the learned policy $\pi_\theta$:
\begin{equation}
    d_t=||w^*_t-w_t||.
\end{equation}
We define function $f$ as:
\begin{equation}
    f(d_t)=C(\psi_{t:t+1})+C(\psi^*_{t+1})-C(\psi^*_t),
\end{equation}
where $\psi_{t:t+1}$ denote the path between the current robot position $p_t$ and selected waypoint $w_t$, thus $C(\psi_{t:t+1})+C(\psi^*_{t+1})$ is the length of the optimal exploration path if the robot selects $w_t$ (remind that ($C(\psi^*_t)$) is the length of the optimal exploration path if the robot chooses the optimal waypoint $w^*_t$). Consequently, $f\geq0$ measures the increase of optimal exploration path length resulting from the current waypoint decision. We note that:
\begin{equation}
    \sum_{t=1}^nf(d_t)=C(\psi^*_{n+1})+\sum_{t=1}^n[C(\psi_{1:t})]-C(\psi^*_1),
\end{equation}
where $n$ denotes the step when the exploration is completed, $C(\psi^*_{n+1})=0$ since the robot does not need to move anymore, and $\sum_{t=1}^nC(\psi_{1:t})=C(\psi_n)$ is the length of the full executed exploration path. As a result, we get:
\begin{equation}
    \sum_{t=1}^nf(d_t)=C(\psi_n)-C(\psi^*_1),
\end{equation}
where the right-hand side is equivalent to the exploration objective defined in Equation~\ref{eq: objective}, as the length of the optimal path $C(\psi^*_1)$ is a constant. When $f$ is 
monotonically increasing, minimizing $C(\psi^*_1)$ is align with minimizing $\sum_{t=1}^nd_t$. Although the monotonicity assumption of $f$ does not hold in special scenarios, where multiple equivalent optimal expert paths exist, such scenarios happen rarely in practice. Moreover, remaining close to the expert still optimizes the objective in these scenarios. Therefore, we believe it is fair to conclude that \textbf{minimizing $\mathbb{E_{\pi_{\theta}}}[\sum_{t=1}^nd_t]$ is a consistent surrogate of the exploration objective in general}.
As a result, we define the reward $r_t$ as:
\begin{equation}
    r_t=-\frac{e^{d_t/2d_n}-1}{e-1}, r_t\in[-1,0]
\end{equation}
where $d_n$ is the neighboring threshold to rescale $d_t$ between $[0,1]$. The exponential term further encourages the policy to remain close to the expert’s action.
Compared to prior DRL-based methods that rely on carefully tuned, biased rewards~\cite{niroui2019deep,chen2019self,xu2022explore,chen2020autonomous,cao2023ariadne,zhu2024maexp}, our privileged expert reward is both simpler and consistent with the true objective. 

Regarding the computing efficiency, although solving TSP is usually computationally expensive, our algorithm only conducts TSP over sampled viewpoints. In our training dataset, environments usually could be covered by fewer than 100 viewpoints. Besides, in practice, we do not seek the optimal solution, but just a fair solution, so we run OR-Tools~\cite{ortools} to output the initial solution without fine optimization. As a result, during our training, the TSP computation is less than a second. That being said, although we find that models trained in relatively small environments are sufficient to generalize in very large scale (no matter how large the environment is, the robot only needs to care about the nearby global reference), our current implementation of the expert path planning may not be efficient if directly doing training in very large environments (e.g., requiring more than 1000 viewpoints to cover). But it should be easy to solve by implementing the algorithm to leverage our global graph and conduct hierarchical planning.

\subsubsection{Training Details}
We train our policy network with soft actor critic~\cite{haarnoja2018soft,christodoulou2019soft}. The goal of SAC is to balance the trade-off
between maximizing returns and policy entropy:
\begin{equation}
\label{eq: training objective}
    \pi^*_\theta= \mathop{\rm {argmax}} \mathbb{E}[\sum^T_{t=0}\gamma^t(r_t+\alpha \mathcal{H}(\pi(.|o_t)))],
\end{equation}
where $\mathcal{H}$ denotes the entropy and $\alpha$ is a temperature parameter that tunes the importance of the entropy term versus the return.
SAC trains critic networks $\phi$ to estimate the soft action-state values $Q_{\phi}(o_t, a_t)$, using the critic loss: 
\begin{equation}
\begin{aligned}
    & J_Q(\phi)=\mathbb{E}_{o_t,a_t}[\frac{1}{2}(Q_\phi(o_t,a_t)-(r_t+\gamma \mathbb{E}_{o_{t+1}}[V(o_{t+1})]))^2],\\
    & V(o_t) = \mathbb{E}_{a_t}[Q_\phi(o_t,a_t)]-\alpha {\rm log}(\pi(\cdot|o_t)).
\end{aligned}
\end{equation}
The policy network $\theta$ is trained to output a policy that maximizes the expected state-action value, where the policy loss reads: 
\begin{equation}
J_\pi(\theta)=\mathbb{E}_{o_t,a_t}[\alpha {\rm log}(\pi_\theta(a_t|o_t))-Q_\phi(o_t,a_t)].
\end{equation}
The temperature parameter is auto-tuned during training and the temperature loss is calculated as:
\begin{equation}
   J(\alpha)=\mathbb{E}_{a_t}[-\alpha({\rm log}\pi_t(a_t|o_t)+\overline{\mathcal{H}})],
\end{equation}
where $\overline{\mathcal{H}}$ denotes a target entropy~\cite{haarnoja2018soft,christodoulou2019soft}.

We train our model using environments created by $4000$ randomly generated dungeon maps, each contains $250\times 250$ pixels. Assuming map resolution $\Delta_{map}=0.4m$, we set the sensor range $d_{sensor}=20m$, the utility range $d_{utility}=0.8\times d_{sensor}$. For graph-related parameters, we set the node resolution $\Delta_{node}=4m$, neighboring threshold $d_n=2\sqrt{2}\Delta_{node}$, making each local node has maximum $25$ neighbors. We set the local map size $d_{local}=40m$, the max number of local nodes in a community rounds $\frac{1}{10}(d_{local}/\Delta_{node})^2$. For training-related parameters, we set the feature dimension $d_f=128$, the maximum episode step $200$, the replay buffer size $1\times10^5$, the minimal buffer size $1\times10^4$, the batch size $128$, the learning rate $1\times10^{-5}$ for the policy and critic networks, $1\times10^{-4}$ for the temperature tuning, and the reward discount $\gamma=0.95$. We parallel the data collection in $16$ environments using~\cite{moritz2018ray}. We conduct the training on a desktop equipped with an AMD Ryzen 7 5700X CPU
and an NVIDIA GeForce RTX 4080 GPU. The training takes approximately 16 hours to converge, using around 16 GB of RAM and 2GB of VRAM. We will release the full code upon acceptance.

\subsection{Terrain Segmentation}
We develop a terrain segmentation tool based on the terrain analysis module in~\cite{cao2022autonomous} and Octomap~\cite{hornung13auro}. Two separate OctoMaps are then constructed from the free and obstacle clouds, respectively. At each decision step, we generate the planning map by projecting both Octomaps onto a 2D plane, where obstacle voxels take precedence over free voxels. Since Octomap is able to remove some dynamic obstacles through a raycast mechanism, in practice, we also use the free clouds to help detect and remove dynamic obstacles. Note that this tool is just for quick validation of HEADER, and any other better-developed terrain segmentation approach should also be compatible with HEADER seamlessly. 

\section{Experiments}

We conduct a series of experiments to validate both the effectiveness of our key components and the overall performance of HEADER. We begin with ablation studies in simplified simulation environments, enabling us to evaluate HEADER's average performance across hundreds of environments. We then compare HEADER against state-of-the-art exploration planners in large-scale, realistic Gazebo simulations, covering both indoor and outdoor benchmark scenarios of up to $330m \times 250m$. Finally, we deploy HEADER on physical hardware in three real-world environments, again including both indoor and outdoor scenarios of up to $300m \times 230m$. Note that we use an identical trained model in all the experiments.

\subsection{Ablation Study}
\label{sec: ablation}

We test HEADER in unseen $100$ environments generated by training-like dungeon maps and compare the results against: (1) \textbf{ARiADNE}: It is an attention-based planner proposed in~\cite{cao2023ariadne}, which could be seen as the base policy without the hierarchical graph, global and local guideposts, and privileged expert reward. (2) \textbf{ARiADNE*}: It is a ARiADNE's variant with the local guidepost. (3) \textbf{NG}: It is a HEADER's variant that does not contain hierarchical graph and global guideposts. (4) \textbf{RS}: It is a HEADER's variant trained by reward shaping proposed in~\cite{cao2023ariadne} instead of our privileged expert reward. (5) \textbf{BC}: It is a HEADER's variant trained by behavior cloning over the privileged action. (6) \textbf{TARE Local}: It is a conventional algorithm proposed in~\cite{cao2021tare}, which could be seen as a non-privileged version of our expert planner. (7) \textbf{Optimal}: It is the privileged expert planner that we develop as the training demonstration, which outputs near-optimal paths. The sensor and graph setups are identical to those used in the training.

\begin{table*}[t]
\caption{
\textbf{Comparisons in testing dataset (identical 100 scenarios for each method).}}
\label{table:1}
\begin{center}
\begin{tabular}{c|ccccccc|c}
\toprule
& ARiADNE & ARiADNE* & NG & RS & BC & HEADER & TARE Local & Optimal\\
\midrule
Distance (m) & 606($\pm89$) &579($\pm82$) & 552($\pm68$) & 627($\pm83$) & 661($\pm91$) & \textbf{529}($\pm67$) & 558($\pm70$) & 502($\pm61$)  \\
Gap to optimal & 20.7\% &15.3\% & 10.0\% & 25.0\% & 31.7\% & \textbf{5.3\%} & 11.1\% & 0\%\\
\bottomrule
\end{tabular}
\end{center}
\end{table*}

\begin{figure}[t]
    \centering
    \includegraphics[width=0.8\linewidth]{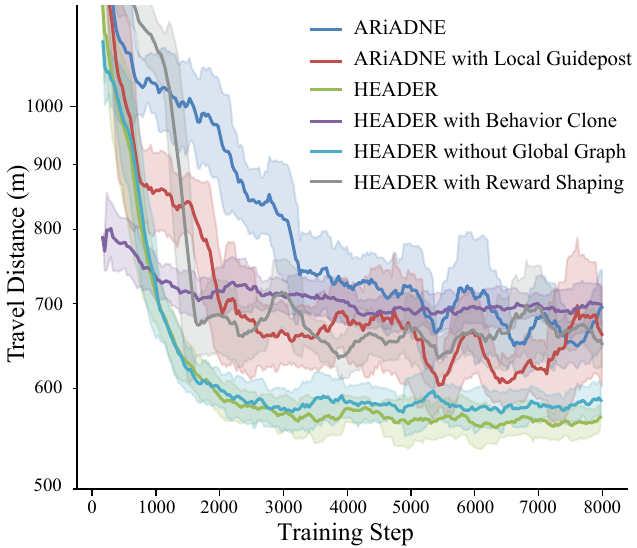}
    \caption{Ablation results of HEADER and ARiADNE in terms of travel distance to complete exploration during training (lower is better).}
    \label{fig:ablation}
\end{figure}

\begin{figure*}[b]
\centering
\subfloat[Tunnel]{
    \includegraphics[width=0.32\textwidth]{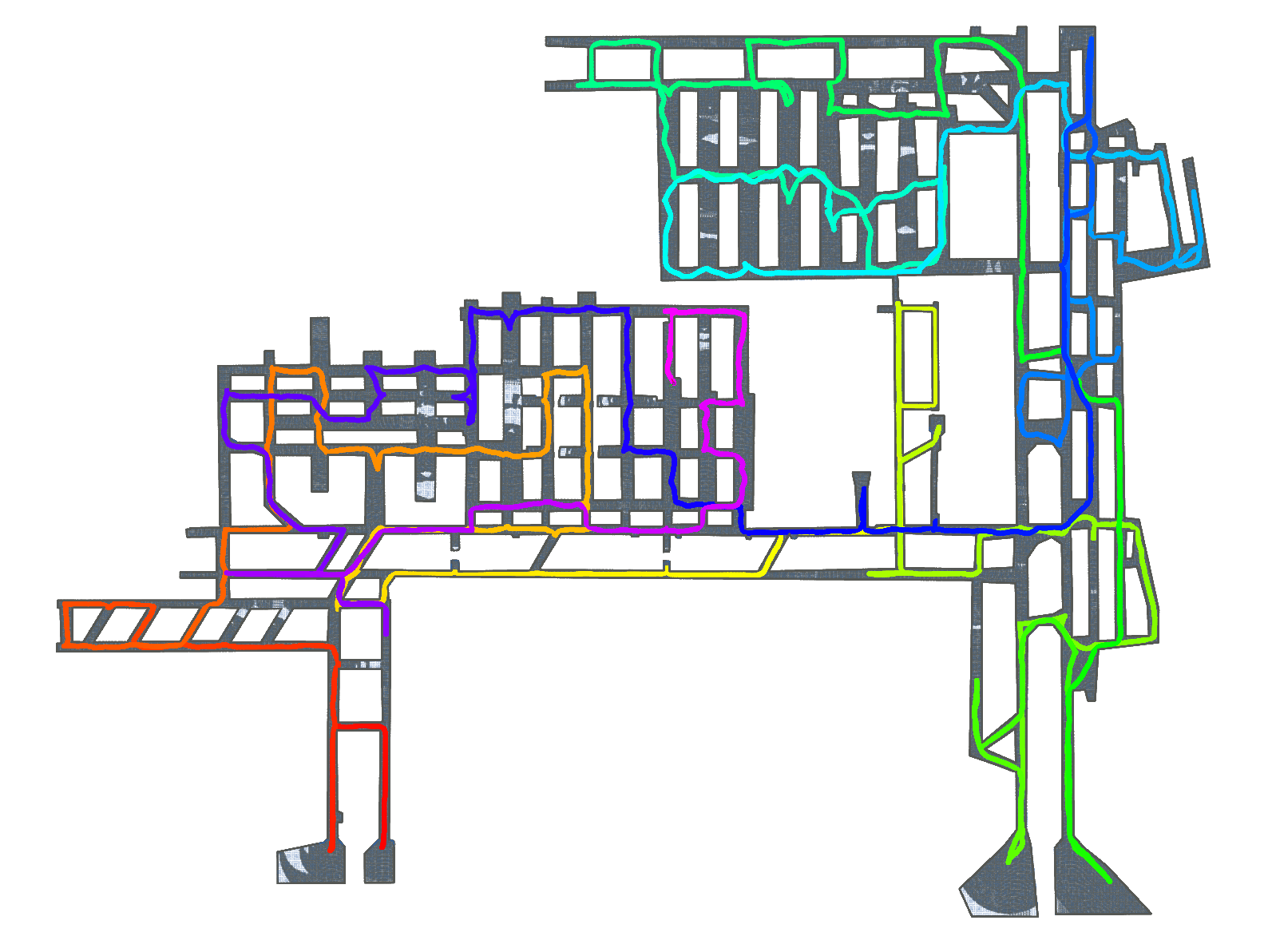}
}%
\subfloat[Campus]{
    \includegraphics[width=0.20\textwidth]{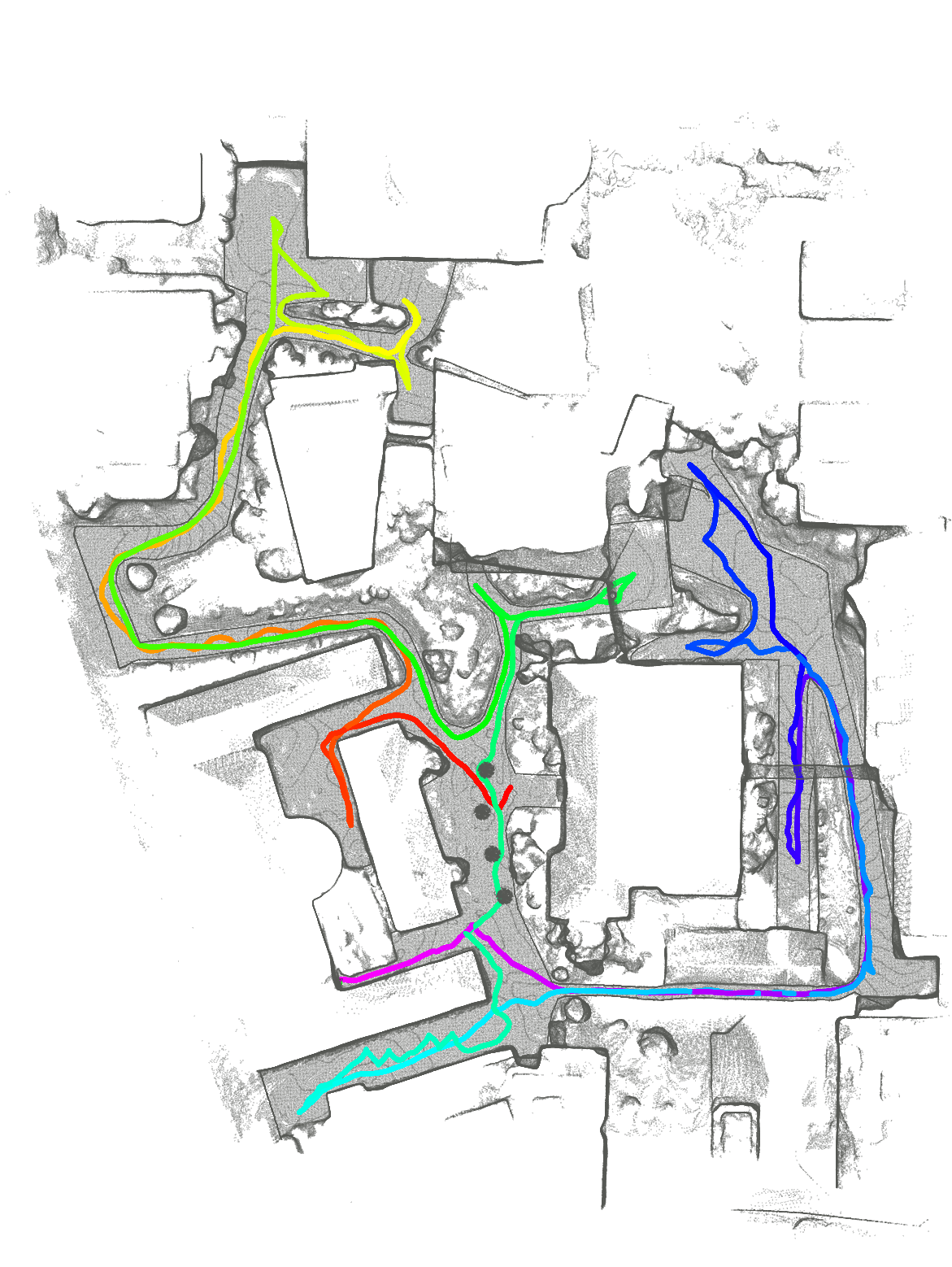}
}%
\subfloat[Forest]{
    \includegraphics[width=0.24\textwidth]{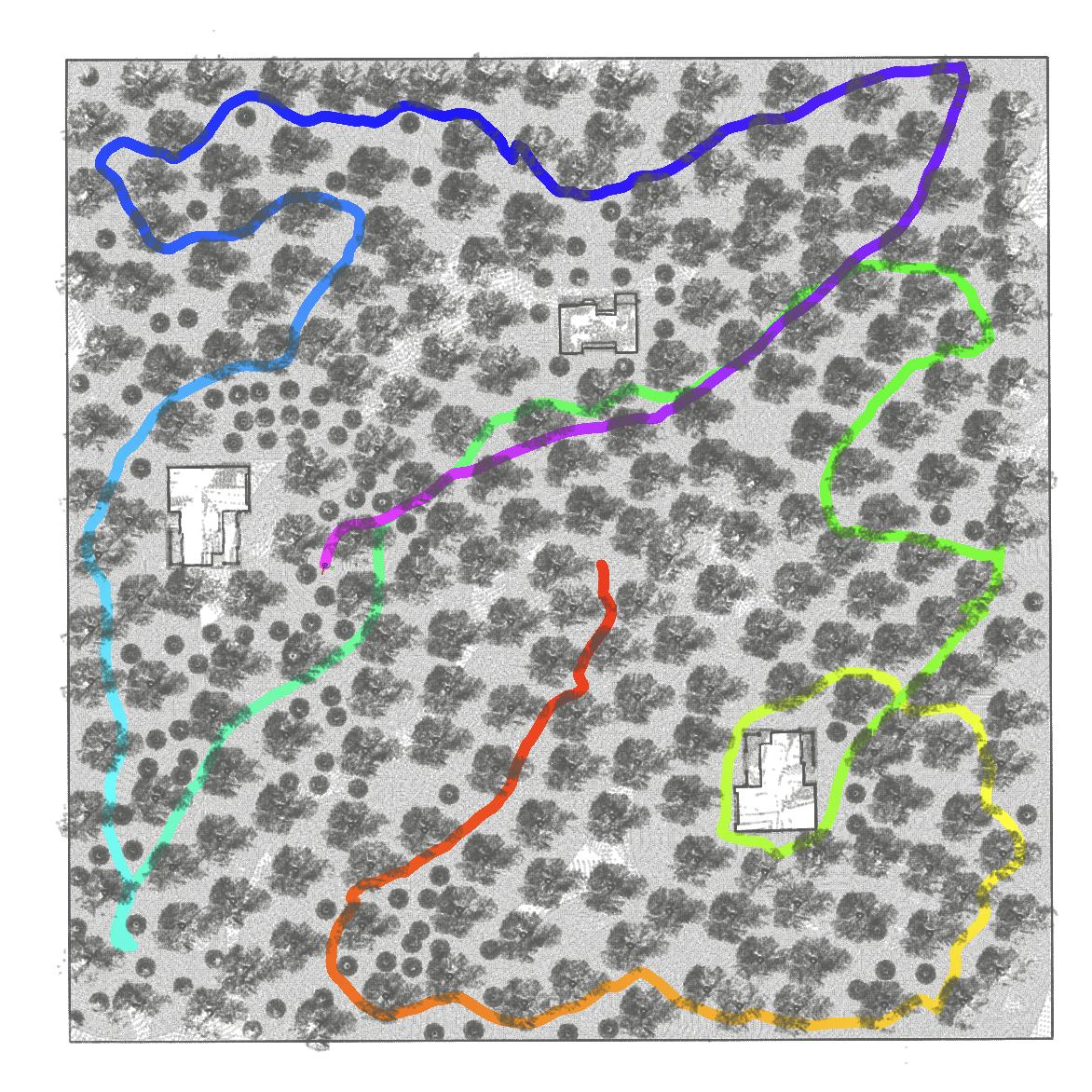}
}%
\subfloat[Indoor]{
    \includegraphics[width=0.18\textwidth]{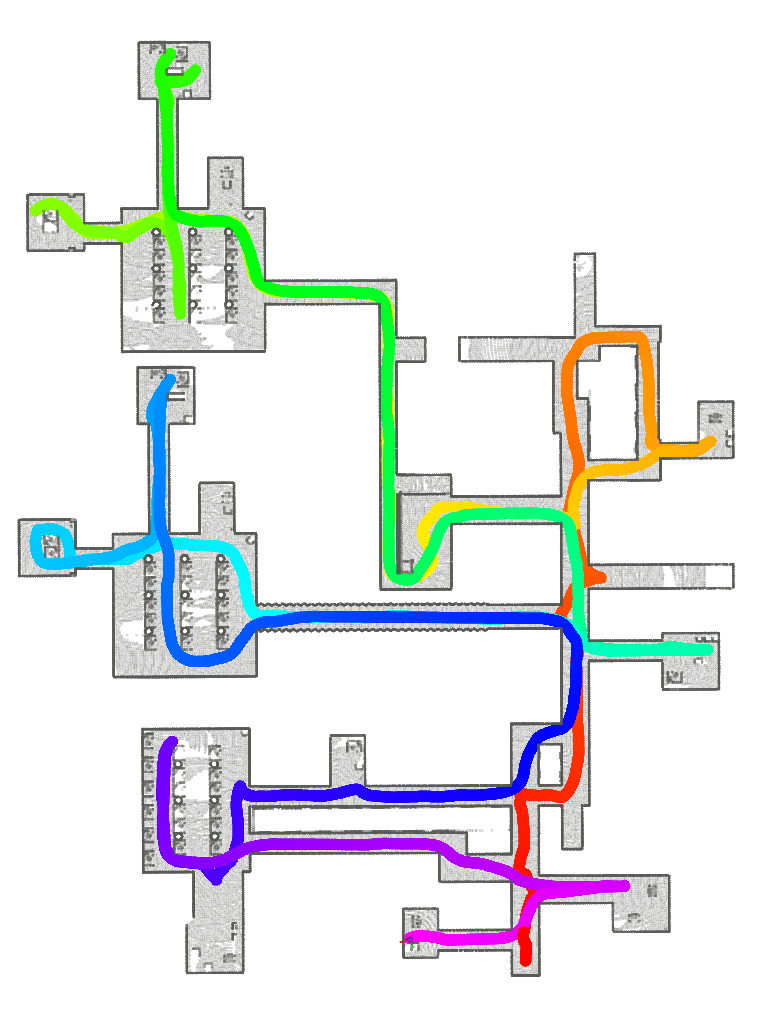}

}
\caption{Demonstration of the exploration trajectories output by HEADER in both indoor and outdoor Gazebo simulations. The Trajectory is color-coded to represent the robot's movement over time.}
\label{fig:ros_sim}
\end{figure*}

\begin{figure*}[t]
\centering
\subfloat[Tunnel\label{fig:tunnel}]{
    \includegraphics[width=0.24\textwidth]{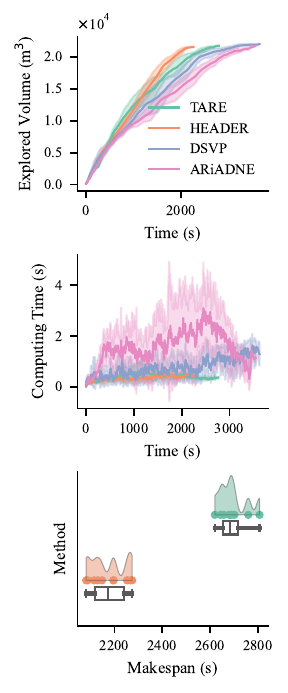}
}%
\subfloat[Campus\label{fig:campus}]{
    \includegraphics[width=0.24\textwidth]{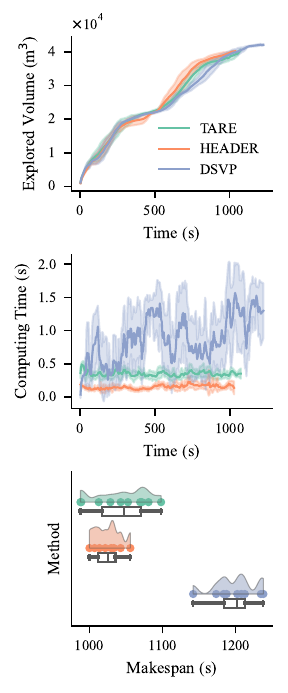}
}%
\subfloat[Forest\label{fig:forest}]{
    \includegraphics[width=0.24\textwidth]{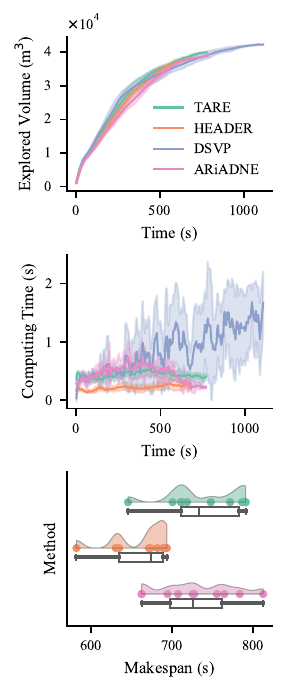}
}%
\subfloat[Indoor\label{fig:indoor}]{
    \includegraphics[width=0.24\textwidth]{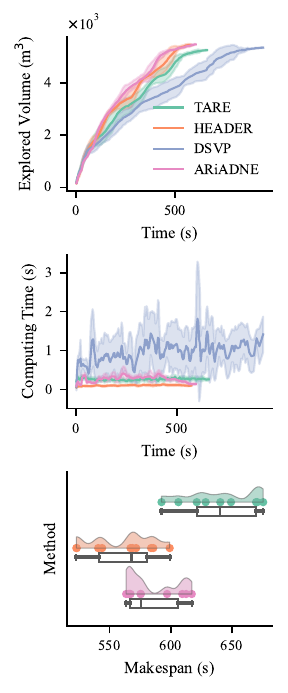}
}
\caption{Performance comparison between HEADER and conventional baselines across 10 runs per method in each environment. The first row shows the explored volume as a function of time, reflecting exploration efficiency. The second row presents the computation time per decision, indicating computational cost during exploration. The third row illustrates the distribution of makespan values of all runs (total time needed to fully complete the exploration task).
Overall, HEADER achieves consistently higher exploration efficiency and lower computation time, and its makespan is statistically lower than that of all baselines.}
\label{fig:ros_results}
\end{figure*}

The comparison results are shown in Table~\ref{table:1}. We also provide the training curves in Fig.~\ref{fig:ablation}. Through analyzing these ablation results, we draw the following conclusions on HEADER's advantages:

\subsubsection{Privileged expert reward is better than reward shaping}
We note that the reward design in RS is common in existing DRL-based works, which consists of a reward to encourage observing unexplored areas, a punishment for the travel distance, and a final reward upon completing the exploration. In environments with varying structures, the weights of these components must be carefully tuned to align the training objective (Equation~\ref{eq: training objective}) with the true exploration objective (Equation~\ref{eq: objective}). Although we have tried our best to tune RS, HEADER trained by the privileged expert reward still significantly outperforms RS ($20\%$ shorter travel distance to complete exploration). RS even performs worse than ARiADNE, which is also trained by the same reward structure, implying that reward shaping is not effective for modeling our new hierarchical observation. Besides, we note that ARiADNE can outperform TARE Local in the dataset proposed in~\cite{chen2019self} but is surpassed by TARE Local in our test dataset. This is because environments in our test dataset are both larger-scale and with more complex structures, where the robot often needs to take long detours to backtrack, making the RS reward inevitably sparse and failing to effectively guide the learning process. In contrast, \textbf{our parameter-free privileged expert reward ensures that the training objective remains strictly aligned with the true exploration goal}, while providing dense feedback in all types of environments. We note that the only difference between RS and HEADER is whether to use privileged expert reward.

\subsubsection{Guideposts enable efficient hierarchical planning}
The superior performance of ARiADNE* over ARiADNE demonstrates the effectiveness of the local guidepost, while the superior performance of HEADER over NG highlights the contribution of the global guidepost. Intuitively, the local guidepost enables the model to capture all unexplored areas more effectively, even when some frontiers are far away. Upon closer inspection, we observe that this is particularly beneficial in scenarios with long corridors, where the original ARiADNE struggles to account for distant utilities.
We also tried assigning the local guidepost only to the nearest utility node, rather than to all utility nodes. However, this results in degraded performance, likely due to the model overfitting to chase the nearest frontier.

The global guidepost further enhances overall performance. Reasoning about long-range dependencies at high resolution is challenging for a learning-based method. Therefore, the performance gain from incorporating global paths into local decision-making is expected. A key advantage of our guidepost design lies in its ability to adaptively and cooperatively integrate global and local decisions.
In contrast, many existing approaches~\cite{chen2019self,dang2020graph,huang2023fael} only perform global path planning when no nearby frontiers are detected, which often results in myopic local decisions. Other methods~\cite{cao2021tare,long2024hphs} enforce strict connections between local paths and the global path, potentially limiting local flexibility.
In our design, \textbf{the global guidepost serves as a soft reference, enabling the model to consider both global and local signals jointly while retaining the flexibility} to ignore the global path when it deems it suboptimal. As a result, our method achieves a better balance between long-range planning and local adaptability.

\subsubsection{HEADER can reason about the unknown area}
A known limitation of learning from demonstration is that the learned policy typically cannot surpass the performance of the expert.
Note that TARE Local and the privileged expert planner share the same path planning algorithm, and the only difference lies in whether the planner has access to the ground-truth information of the unknown areas. Therefore, if HEADER makes decisions solely based on explored areas, it can never outperform TARE Local.

However, our results show that HEADER significantly outperforms TARE Local in finding shorter exploration paths, achieving half the performance gap to the optimal solution. Given that it is the observability of the unknown areas that causes the performance gap between TARE Local and the privileged expert, HEADER's superior performance should be attributed to its ability to implicitly predict some unknown areas during decision-making.

Moreover, an interesting perspective is that the $\sim50\%$ reduction in the gap to the optimal may imply that \textbf{HEADER achieves approximately $50\%$ accuracy in predicting unknown areas that are key to exploration decisions} (at least in the testing dataset). It does not mean HEADER can predict the whole unknown area, but some areas that may significantly influence the exploration decisions. For example, HEADER may predict that two partially observed corridors are likely to connect. While TARE must plan a detour to visit both sides of the corridor separately, HEADER may choose to directly pass through like the privileged planner.
Therefore, we believe our results support the hypothesis that HEADER can effectively reason about the unknown areas based on its experience.

\begin{table}[t]
\caption{
\textbf{Comparison results in the indoor benchmark.}}
\label{table:indoor}
\begin{center}
\begin{tabular}{c|ccc}
\toprule
& Distance ($m$) & Time ($s$) & Computing ($s$) \\
\midrule
HPHS & 1512($\pm157$) & 889($\pm179$) & 0.17($\pm0.06$) \\
FAEL & & Failed & \\
GBP & 1426($\pm114$) & 1079($\pm73$) & 0.67($\pm0.20$)\\
DSVP & 1371($\pm164$) & 820($\pm106$) & 0.96($\pm0.65$) \\
TARE & 1195($\pm76$) & 651($\pm43$) & 0.26($\pm0.06$)\\
ARiADNE & 1005($\pm39$) & 585($\pm21$) & 0.27($\pm0.11$)\\
\midrule
HEADER & \textbf{1000($\pm39$)} & \textbf{561($\pm25$)} & \textbf{0.11($\pm0.02$)} \\
\bottomrule
\end{tabular}
\end{center}
\end{table}

\begin{table}[t]
\caption{
\textbf{Comparison results in the forest benchmark.}}
\label{table:forest}
\begin{center}
\begin{tabular}{c|ccc}
\toprule
& Distance ($m$) & Time ($s$) & Computing ($s$) \\
\midrule
HPHS & 2497($\pm97$) & 1203($\pm58$) & 0.19($\pm0.12$) \\
FAEL & & Failed & \\
GBP & 1905($\pm119$) & 2559($\pm109$)) & 0.66($\pm0.22$))\\
DSVP & 2029($\pm78$) & 1071($\pm39$) & 0.85($\pm0.64$) \\
TARE & 1398($\pm79$) & 737($\pm46$) & 0.44($\pm0.09$)\\
ARiADNE & 1216($\pm91$) & 729($\pm47$)& 0.47($\pm0.24$) \\
\midrule
HEADER & \textbf{1145($\pm70$)} & \textbf{672($\pm47$)} & \textbf{0.21($\pm0.08$)} \\
\bottomrule
\end{tabular}
\end{center}
\end{table}

\begin{table}[t]
\caption{
\textbf{Comparison results in the campus benchmark.}}
\label{table:campus}
\begin{center}
\begin{tabular}{c|ccc}
\toprule
& Distance ($m$) & Time ($s$) & Computing ($s$) \\
\midrule
HPHS & & Failed &  \\
FAEL & & Failed & \\
GBP & & Failed & \\
DSVP & 2291($\pm107$) & 1188($\pm40$) & 0.87($\pm0.55$) \\
TARE & 2038($\pm64$) & 1043($\pm36$) & 0.35($\pm0.08$)\\
ARiADNE &  & Failed & \\
\midrule
HEADER & \textbf{1894($\pm67$)} & \textbf{1029($\pm58$)} & \textbf{0.14($\pm0.06$)} \\
\bottomrule
\end{tabular}
\end{center}
\end{table}

\begin{table}[t]
\caption{
\textbf{Comparison results in the tunnel benchmark.}}
\label{table:tunnel}
\begin{center}
\begin{tabular}{c|ccc}
\toprule
& Distance ($m$) & Time ($s$) & Computing ($s$) \\
\midrule
HPHS & & Failed &  \\
FAEL & & Failed & \\
GBP & & Failed & \\
DSVP & 6479($\pm212$) & 3488($\pm123$) & 0.75($\pm0.56$) \\
TARE & 5040($\pm219$) & 2693($\pm126$) & 0.41($\pm0.22$)\\
ARiADNE &  4499($\pm280$) & 3476($\pm416$) & 1.77($\pm1.48$) \\
\midrule
HEADER & \textbf{3814($\pm133$)} & \textbf{2175($\pm69$)} & \textbf{0.34($\pm0.16$)} \\
\bottomrule
\end{tabular}
\end{center}
\end{table}

\subsection{Comparison Analysis}

\begin{figure*}[t]
\centering
\subfloat[TARE]{
    \includegraphics[width=0.4\textwidth]{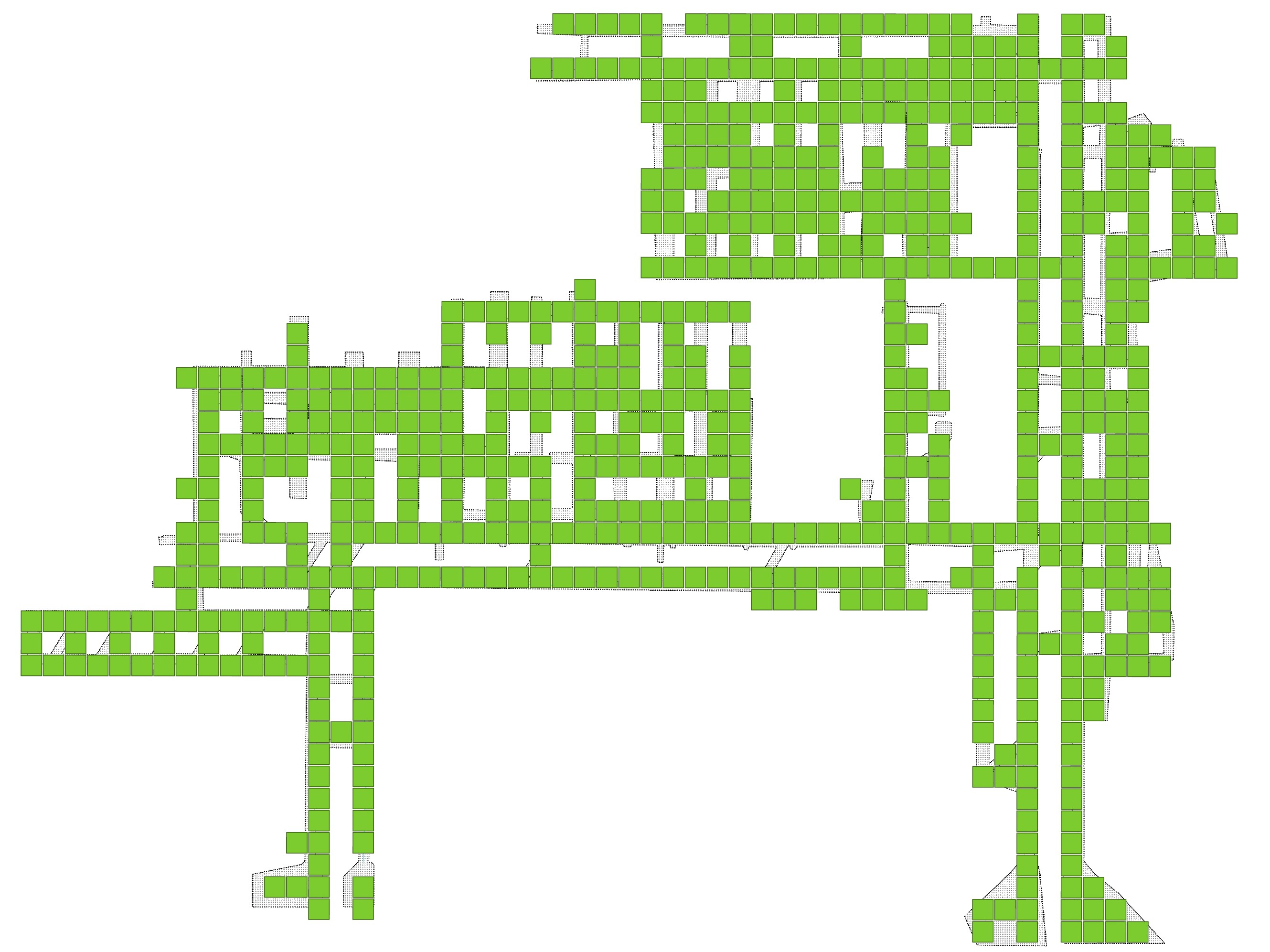}
}%
\hspace{+0.5cm}
\subfloat[HEADER]{
    \includegraphics[width=0.4\textwidth]{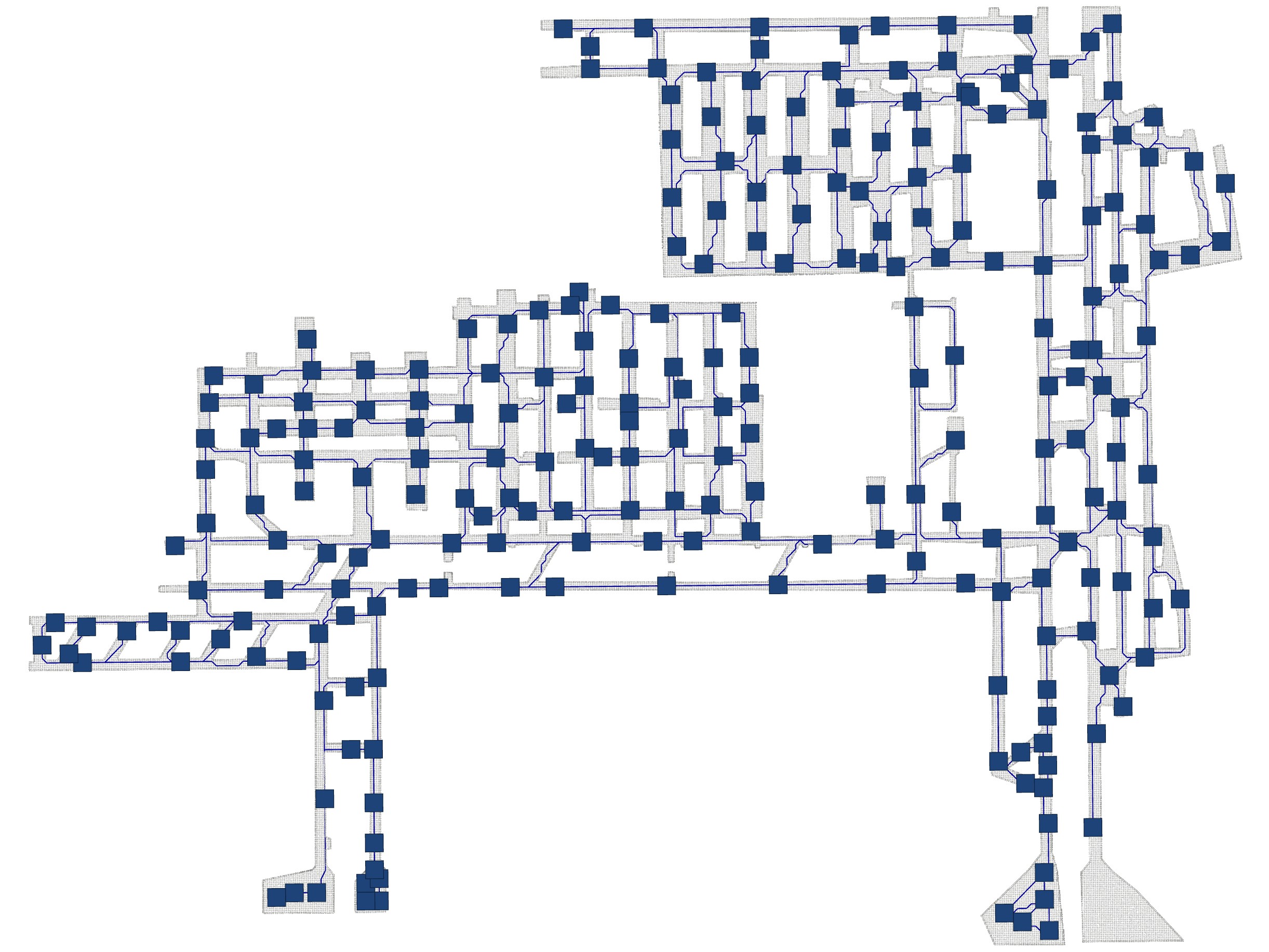}
}%
\caption{Comparison of global representations of HEADER and TARE. TARE creates 697 subspaces as the global representations, while HEADER only needs 250 global nodes to cover the entire environment.}
\label{fig:global_comparison}
\end{figure*}

We integrate HEADER into the robot operating system (ROS) to test and compare it with the state-of-the-art open-sourced exploration planners, including: (1)\textbf{TARE}~\cite{cao2021tare}, (2) \textbf{DSVP}~\cite{zhu2021dsvp}, (3) \textbf{GBP}~\cite{dang2020graph}, (4) \textbf{FAEL}~\cite{huang2023fael}, (5) \textbf{HPHS}~\cite{long2024hphs}, (6) \textbf{ARiADNE}~\cite{cao2024deep}. Note that ARiADNE here is integrated with a graph rarefaction algorithm as in~\cite{cao2024deep} to enhance performance in large-scale exploration, and all graph settings (e.g., node resolution and neighboring threshold) are the same as those in HEADER. We test the exploration performance of HEADER and the baseline planners in four exploration benchmark environments proposed in~\cite{cao2022autonomous}: a $340m\times340m$ campus environment, a $130m\times100m$ indoor corridors environment, a $150m\times150m$ forest environment, and a $330m\times250m$ tunnel network environment. In these benchmarks, the test platform is a four-wheeled differential-drive robot equipped with a 16-channel 3D LiDAR, and the max speed is $2m/s$. We select these benchmarks since they are the largest and most realistic testing environments, to the best of our knowledge. In the campus environment, we block off certain areas, as the current implementation of HEADER cannot handle multilayer environments, where free space may exist above or below occupied regions. All the test runs on an ASUS mini PC with Intel i7-12700H CPUs (which is also used in our hardware validation). Note that GPU is not necessary for HEADER after training.

For different environments, we only tune two parameters for HEADER: the node resolution $\Delta_{node}$, which ranges from $1.2m$ to $2.8m$, and the planning frequency, which ranges from $1Hz$ to $2.5Hz$. The baseline planners typically have more parameters to tune to get the best performance for each environment. We made our best effort to tune them to improve their performance, except TARE and DSVP, which are tuned by their authors.
To enable the usage of HEADER in environments with non-flat ground, we develop a terrain segmentation tool based on a point cloud analysis module proposed in~\cite{cao2022autonomous} and Octomap~\cite{hornung13auro}, which classifies voxels as traversable or occupied. The classified voxels are then projected to formulate an evaluation map as the input of HEADER. 

\begin{figure*}[t]
\centering
\includegraphics[width=0.8\textwidth]{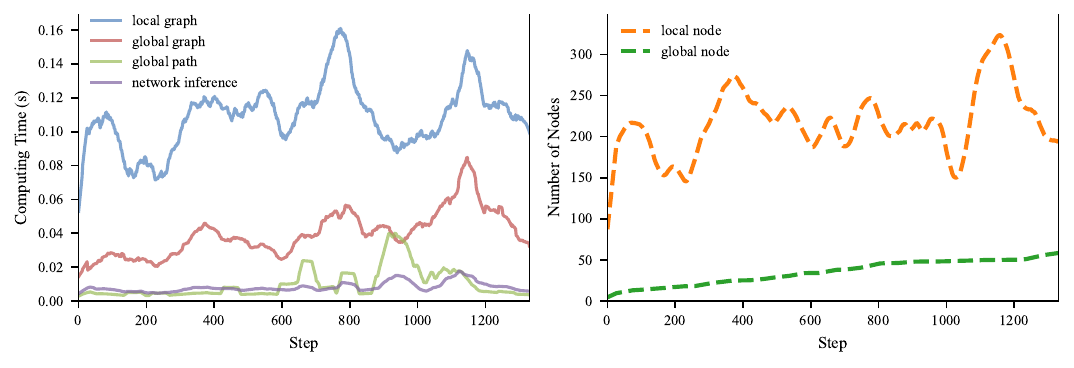}
\caption{Breakdown of HEADER's planning time in the forest environment into local graph update, global graph update, global path planning, and network inference. The trends are shown together with the numbers of local and global nodes to highlight their computational dependencies.}
\label{fig:computing_breakdown}
\end{figure*}

\begin{figure*}[t]
  \centering
  \subfloat[Campus Explored Map\label{fig:cde_map}]{
    \includegraphics[width=0.48\textwidth]{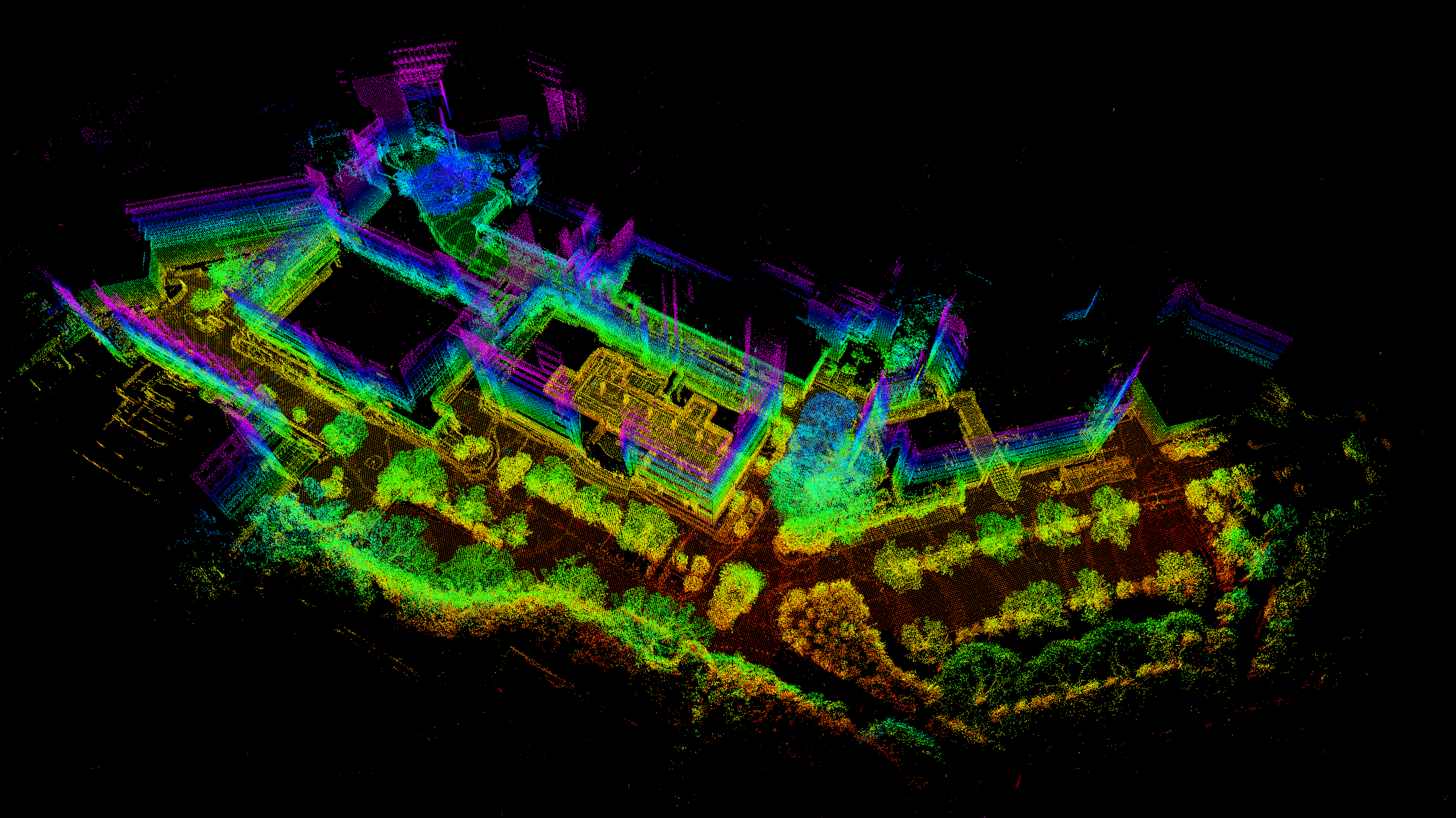}
  }
  \hfill
  \subfloat[Campus Exploration Trajectory\label{fig:cde_traj}]{
    \includegraphics[width=0.48\textwidth]{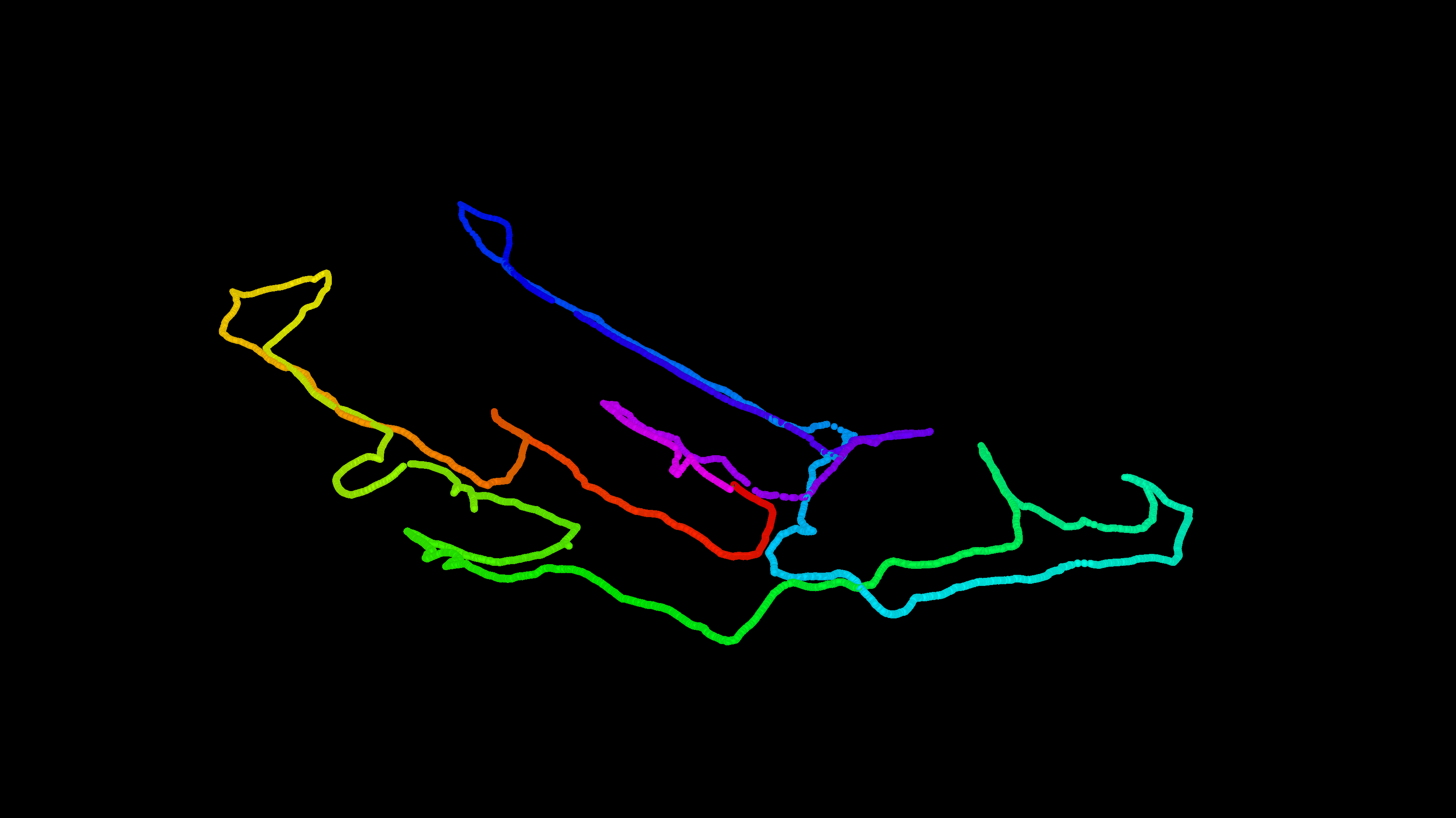}
  }

  \begin{minipage}[t]{0.32\textwidth}
    \vspace{+0.5cm}
    \centering
    \subfloat[Hardware setup\label{fig:robot}]{
      \includegraphics[width=\linewidth]{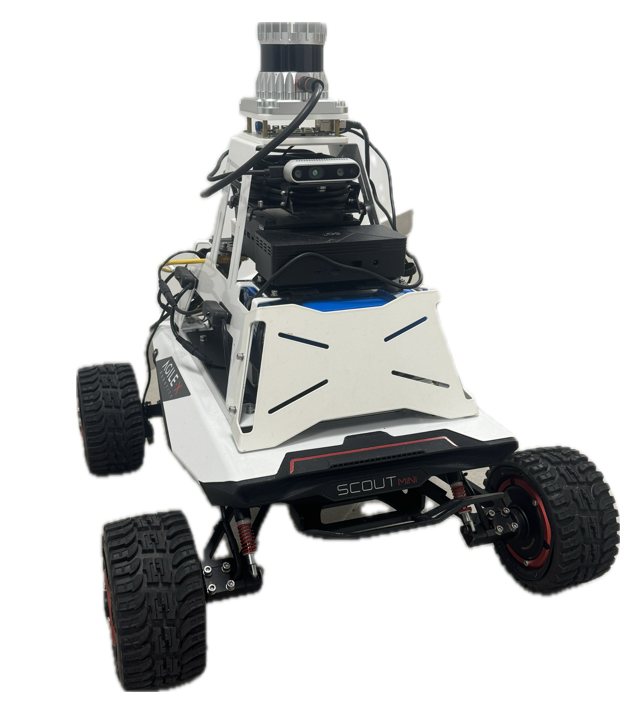}
    }
  \end{minipage}
  \hfill
  \begin{minipage}[t]{0.64\textwidth}
    \centering
    \subfloat[Garden Map\label{fig:garden_map}]{
      \includegraphics[width=0.48\linewidth]{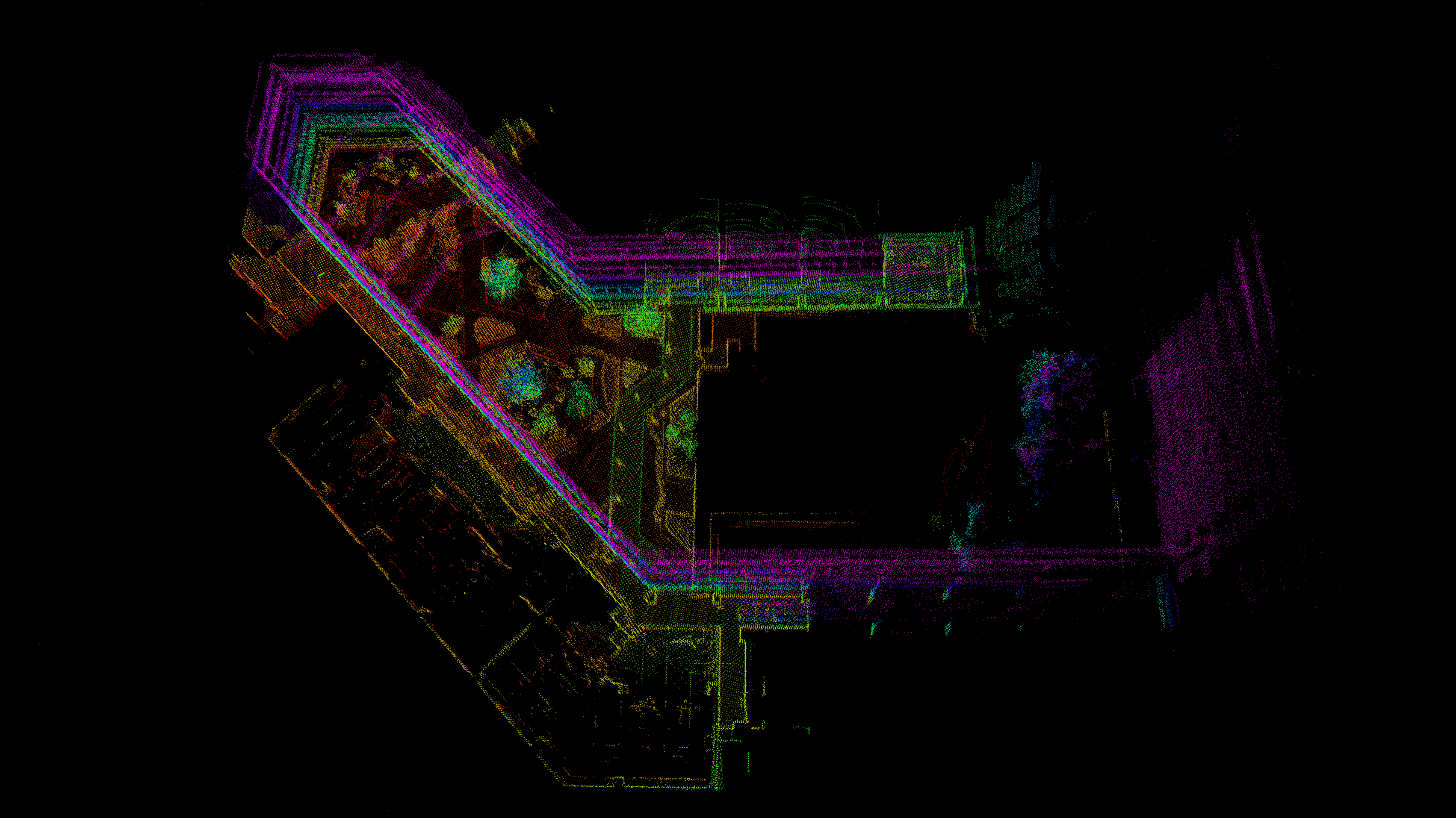}
    }%
    \subfloat[Garden Trajectory\label{fig:garden_traj}]{
      \includegraphics[width=0.48\linewidth]{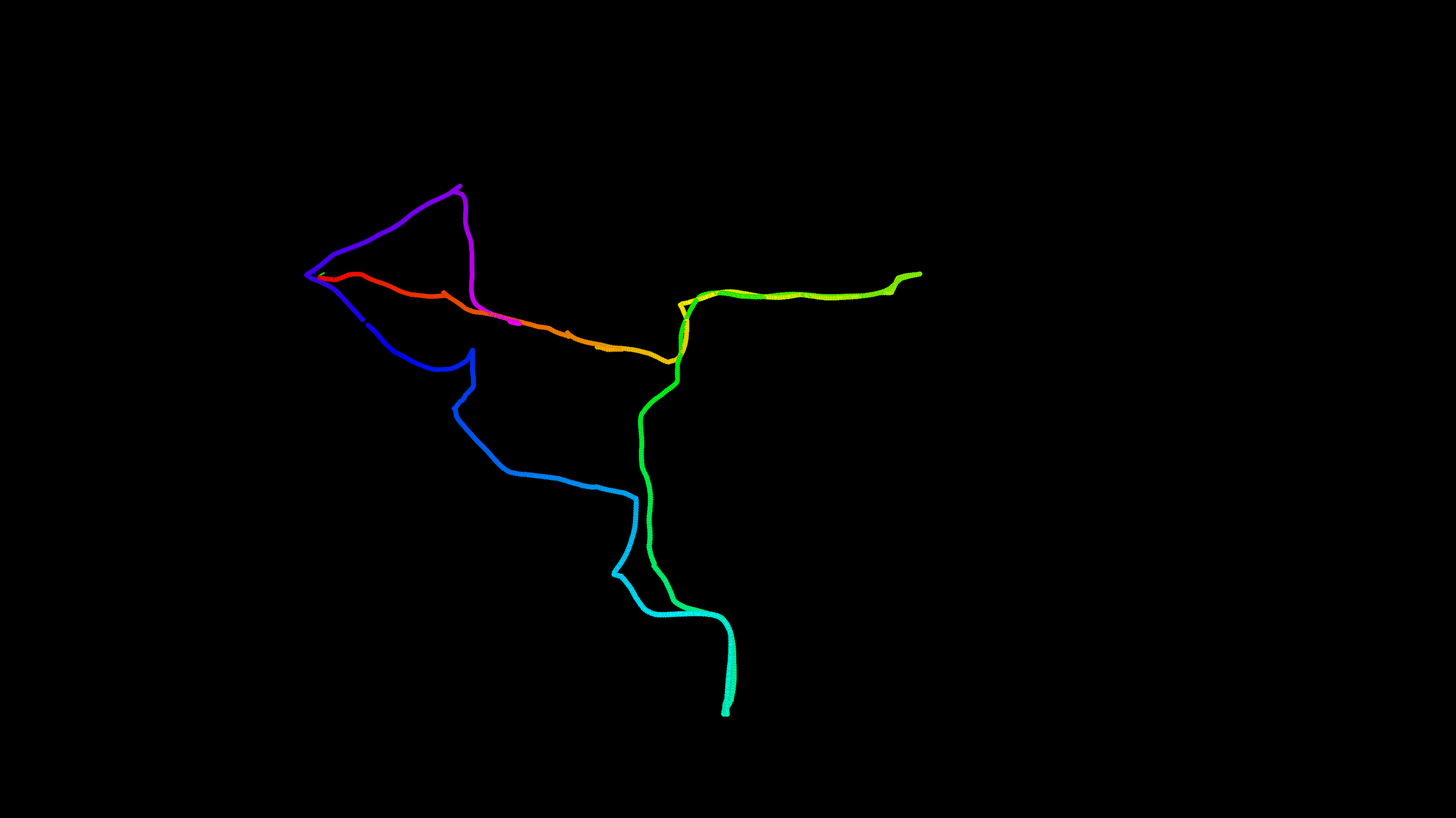}
    }\\[1ex]
    \subfloat[Building Map\label{fig:building_map}]{
      \includegraphics[width=0.48\linewidth]{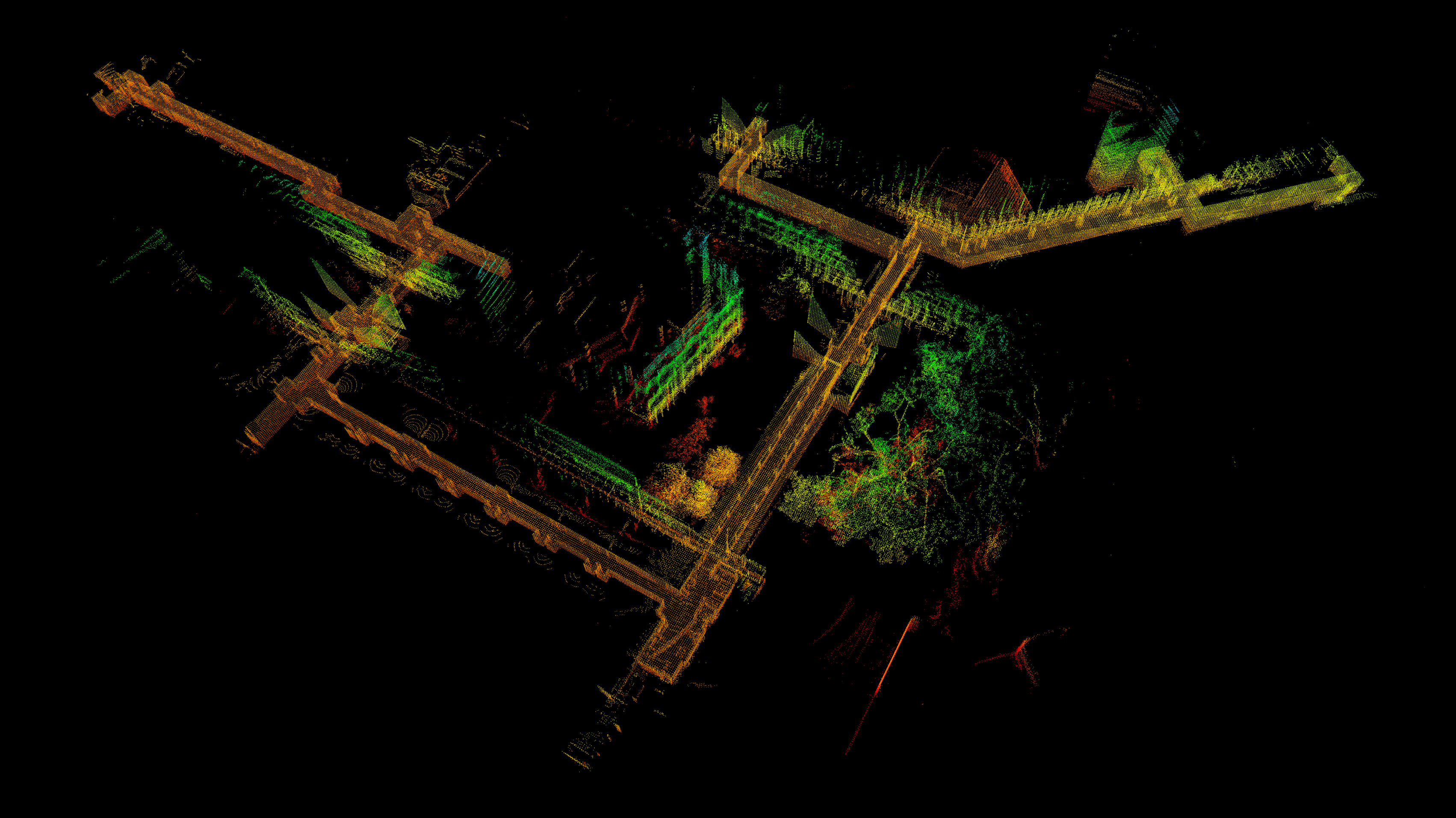}
    }%
    \subfloat[Building Trajectory\label{fig:building_traj}]{
      \includegraphics[width=0.48\linewidth]{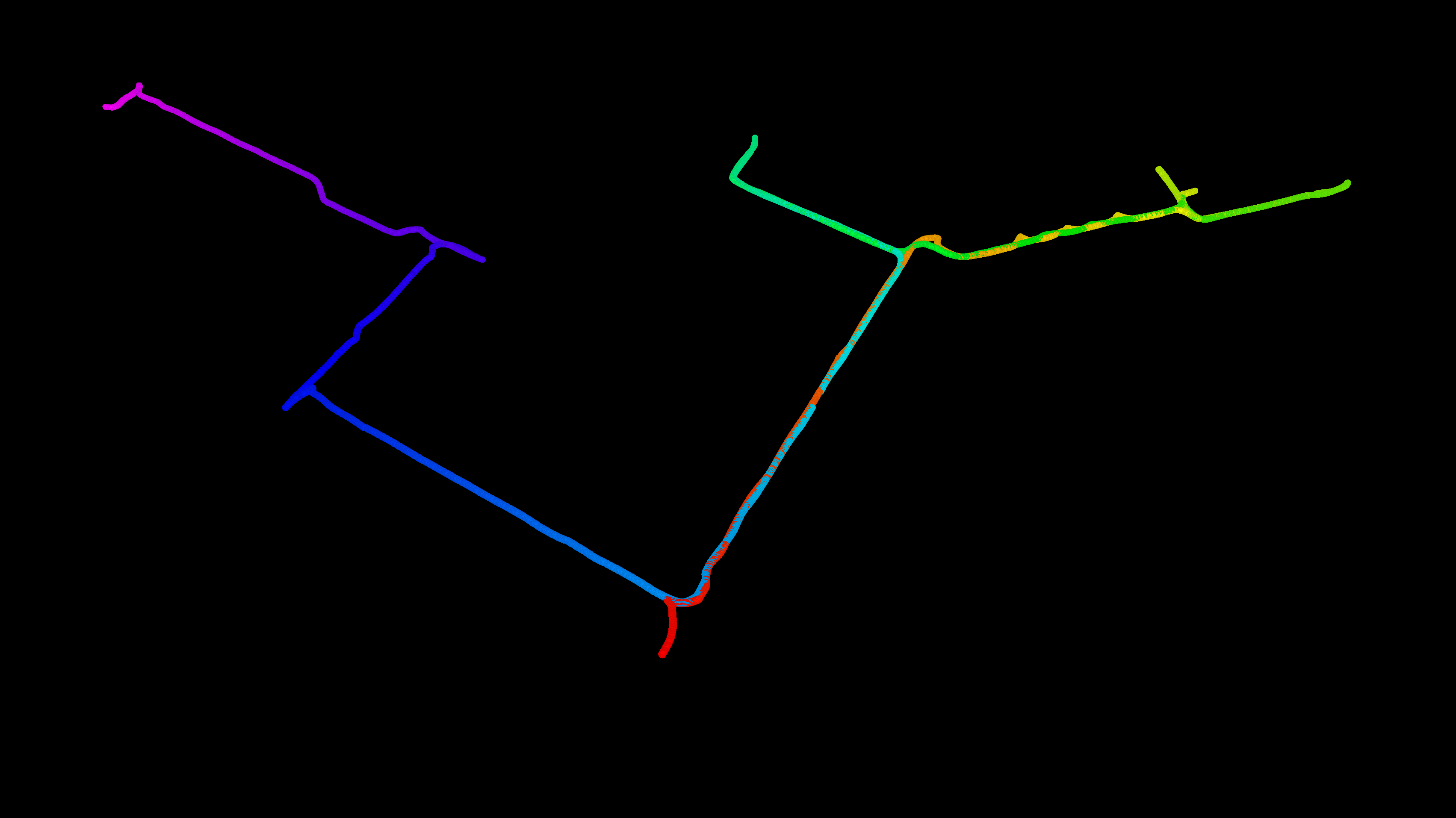}
    }
  \end{minipage}

  \caption{Validations of HEADER in real-world scenarios on a wheeled robot. The trajectory is color-coded to represent the robot's movement over time. The campus environment is around $300\,m \times 230\,m$, garden $75\,m \times 60\,m$, and building $200\,m \times 130\,m$.}
  \label{fig:hardware}
\end{figure*}

We run HEADER, ARiADNE, TARE, DSVP $10$ times and HPHS, FAEL, GBP $5$ times in each environment. The test results are shown in Table~\ref{table:indoor},~\ref{table:forest},~\ref{table:campus}, and~\ref{table:tunnel}. We also analyze and visualize some key results in Fig.~\ref{fig:ros_results}. Among all the baselines, TARE demonstrates the best scalability, making it one of the only two methods capable of completing the exploration task in all four environments. Although HPHS and FAEL report better performance than TARE in smaller or simpler environments~\cite{long2024hphs,huang2023fael}, in our experiments, they fail to complete exploration in some environments when allowed to explore for up to twice the time TARE required. Besides, TARE constantly achieves the shortest travel distance and makespan among all the conventional methods. On the other hand, the learning-based method, ARiADNE, performs well in the indoor benchmark but also suffers from scalability issues in more complex and large-scale environments, which could be seen from its exponentially increased computing time. In the tunnel environment, although ARiADNE achieves a shorter travel distance than TARE, its heavy computation from graph rarefaction leads to a significantly longer makespan, making it clearly outperformed by TARE in terms of overall efficiency.

We note that HEADER consistently outperforms both TARE and ARiADNE across all tested environments. In terms of travel distance, HEADER achieves $16\%$, $18\%$, $6\%$, and $24\%$ reductions over TARE in the indoor, forest, campus, and tunnel environments, respectively.
In terms of average per-step computing time, HEADER demonstrates even greater improvements: $59\%$, $52\%$, $60\%$, and $17\%$ reductions in the above environments respectively, despite being implemented in Python, whereas TARE is implemented in C++. Compared to ARiADNE, HEADER shows comparable performance in the indoor and forest environments. Besides, it successfully completes the exploration task in the campus, where ARiADNE fails. In the tunnel environment, HEADER significantly outperforms ARiADNE, reducing travel distance by $15\%$, makespan by $37\%$, and average computing time by $80\%$:
Upon closer examination, we attribute HEADER's superior performance to three key aspects:

\subsubsection{The learned local decisions better avoid backtrack behaviors}
At the local planning level, as demonstrated in Section~\ref{sec: ablation}, \textbf{HEADER is able to reason about critical unknown areas, enabling it to make decisions that effectively avoid redundant backtracking behavior}. Conventional methods, by contrast, often suffer from inefficient detour exploration paths due to lack of ability to infer information beyond known areas. Moreover, HEADER's advantage over ARiADNE remains evident in high-fidelity simulations and becomes even more significant in complex scenarios.

\subsubsection{The community-based graph is a more efficient global representation}
At the global level, \textbf{our community-based approach enables the construction of a sparser graph representation, offering improved scalability without compromising the coverage of explored areas}. Header's global partitioning depends on the structure of the constructed local graph, allowing the shape of each global partition to be adaptively and flexibly adjusted. For example, in the tunnel environment, TARE creates 697 subspaces to cover the environment, while HEADER only needs 250 global nodes (see Fig.~\ref{fig:global_comparison}). Besides, the computation is typically light since the complexity of our community partition is $\mathcal{O}(n+m)$, where $n$ denotes the number of local nodes and $m$ denotes the number of local edges. Fig.~\ref{fig:computing_breakdown} presents a decomposition of HEADER’s computing time, where the global graph update time shows a linear correlation to the number of local nodes. 

\subsubsection{Joint decisions allow more adaptive exploration behaviors}
Existing methods typically require the robot to strictly follow the planned global path, which is coarse and may be misleading in certain scenarios. In contrast, \textbf{HEADER treats the global path as a reference rather than a constraint, allowing the local planner to make joint decisions}. This enables HEADER to adaptively determine whether to follow the global guidance or deviate from it when necessary. As a result, HEADER demonstrates robustness in scenarios where the global path is suboptimal or potentially misleading for local exploration: Upon closer observation, we find that it is normal that HEADER does not follow the global path when there are frontiers within the local range. It makes sense because our global paths are planned to visit global nodes instead of observing frontiers, but biases at the global planning level will be eliminated through the local network.

\subsection{Hardware Validation}

We validate HEADER on an Agilex Scout-mini wheeled robot equipped with an Ouster OS0-32 LiDAR. We use FastLIO2~\cite{xu2022fast} to get the odometry and mapping. We set the max speed to $1m/s$, sensor range to $8m$, and replanning frequency to $1Hz$ for all of our tests. We run HEADER in three real-world scenarios: a $200m\times130m$ indoor teaching building environment, a $75m\times60m$ outdoor garden environment, and a $300m\times230m$ outdoor campus environment. We set the map resolution $\Delta_{map}=0.4m$, the node resolution $\Delta_{node}=0.8m$ for teaching building and garden environments, and $\Delta_{map}=1m, \Delta_{node}=2m$ for the campus environment. We note that these parameters were all guessed beforehand without tuning during the experiment. Since our implementation has not considered negative obstacles, we manually block stairs in the building and garden environments. Besides, we set an exploration boundary for the campus environments, since it is an open environment. In these real-world tests, HEADER reproduces similar performance as in the simulation: successfully exploring the full environment, keeping low computing time, and achieving high exploration efficiency. We believe this hardware validation further demonstrates HEADER’s generalizability to unseen environments, as well as its robust sim-to-real transferability. 

\section{Conclusion}

In this work, we propose HEADER, a hierarchical learning-based planner for autonomous robot exploration. We first introduce an efficient global representation using community detection. We then develop an attention-based neural network that leverages both global reference paths and local observations to make joint decisions for local movements. Furthermore, we design a privileged expert reward to train the network to produce decisions that benefit long-term efficiency, enabling the model to better reason about critical unknown areas (which can be quantified by improved performance over the expert planner that solely relies on known areas). As a result, HEADER achieves state-of-the-art performance in terms of scalability and exploration efficiency, as demonstrated in both high-fidelity simulations and real-world hardware experiments.

Future work will focus on extending HEADER from single to multi-robot exploration, where robots need to achieve efficient cooperation at both the local and global levels. We are also interested in extending HEADER to handle sensors with a limited field-of-view (e.g., cameras) and 3D action space, where the heading of the robot should be considered.

\bibliographystyle{IEEEtran}
\bibliography{references}

\end{document}